\documentclass[10pt,twocolumn,letterpaper]{article}

\usepackage[pagenumbers]{cvpr} 

\usepackage{times}
\usepackage{epsfig}
\usepackage[accsupp]{axessibility} 
\usepackage{graphicx}
\usepackage{amsmath}
\usepackage{amssymb}
\usepackage{booktabs}
\usepackage{xr-hyper}
\usepackage{caption}
\usepackage{subcaption}

\usepackage[pagebackref,breaklinks,colorlinks]{hyperref}

\usepackage[capitalize]{cleveref}
\crefname{section}{Sec.}{Secs.}
\Crefname{section}{Section}{Sections}
\Crefname{table}{Table}{Tables}
\crefname{table}{Tab.}{Tabs.}

\def\cvprPaperID{6567} 
\def\confName{CVPR}
\def\confYear{2022}

\begin{document}

\title{Grounding Answers for Visual Questions Asked by Visually Impaired People} 

\author{\hspace{-0.1in} Chongyan Chen$^1$, Samreen Anjum$^2$, Danna Gurari$^1$$^,$$^2$ \\
\noindent
{\hspace{-0.2in} \small $~^1$ University of Texas at Austin}
{\small $~^2$ University of Colorado Boulder}
}

\maketitle

\begin{abstract}
  Visual question answering is the task of answering questions about images. We introduce the VizWiz-VQA-Grounding dataset, the first dataset that visually grounds answers to visual questions asked by people with visual impairments.  We analyze our dataset and compare it with five VQA-Grounding datasets to demonstrate what makes it similar and different. We then evaluate the SOTA VQA and VQA-Grounding models and demonstrate that current SOTA algorithms often fail to identify the correct visual evidence where the answer is located. These models regularly struggle when the visual evidence occupies a small fraction of the image, for images that are higher quality, as well as for visual questions that require skills in text recognition.  The dataset, evaluation server, and leaderboard all can be found at the following link: \url{https://vizwiz.org/tasks-and-datasets/answer-grounding-for-vqa/}.
 \end{abstract}
\section{Introduction}

Visual question answering (VQA) is the task of providing a natural language answer to a question about an image.  While most VQA services only return a natural language answer, our work is motivated by the belief that it is also valuable for a VQA service to return the region in the image used to arrive at the answer.  We call this task of locating the relevant visual evidence \emph{answer grounding}.  

\begin{figure}[t!]
\centering
	\includegraphics[width=1\linewidth]{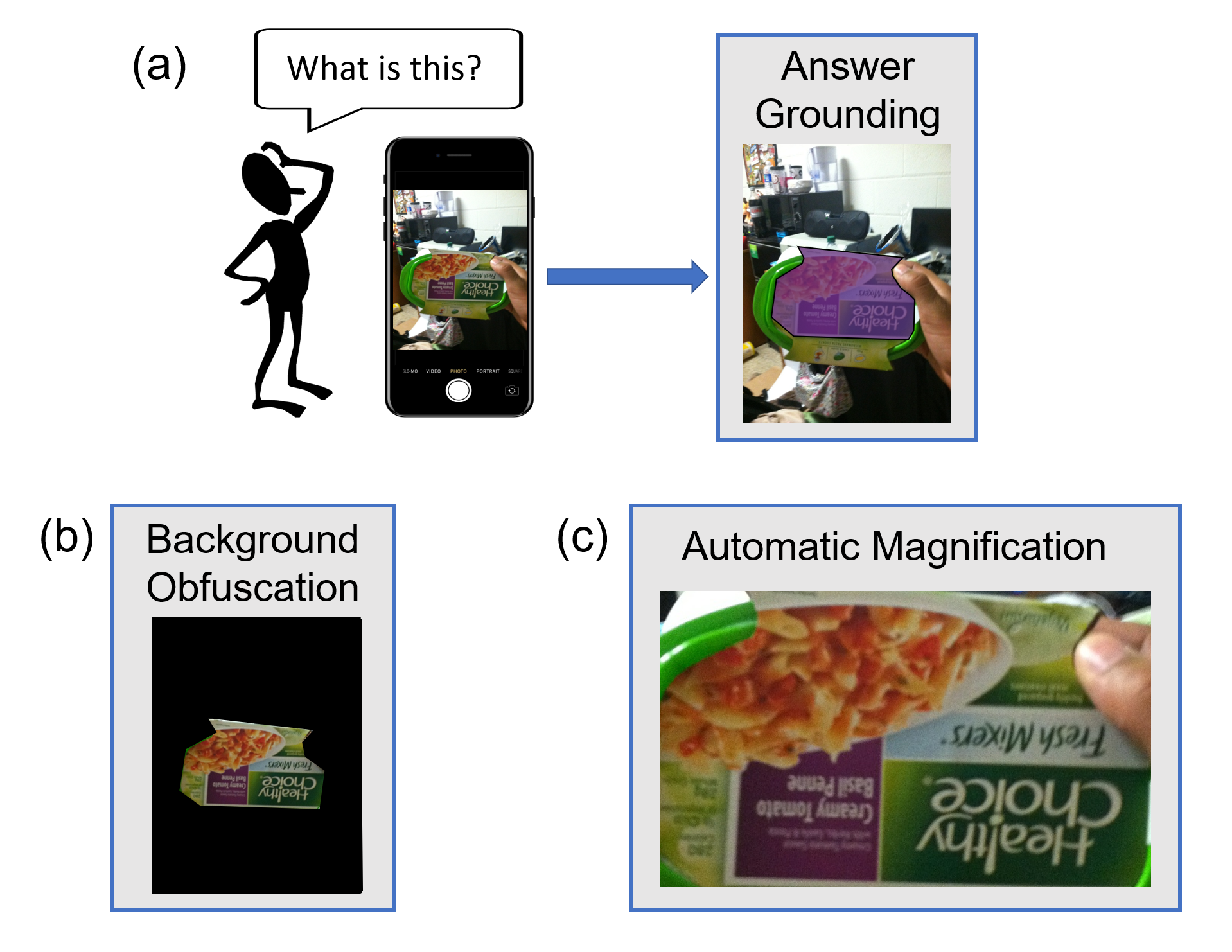} \hfill 
  \vspace{-1.75em}
  \caption{(a) We introduce a new dataset challenge that supports the task of grounding the visual evidence needed to answer visual questions asked by people with vision impairments.  This enables valuable use cases including \textbf{(b)} background obfuscation to limit inadvertent privacy leaks and \textbf{(c)} automatic magnification to expedite low vision users' abilities to answer their questions.}
  \label{fig:use-cases}
\end{figure}

Numerous applications would be possible if answer groundings were provided in response to visual questions. \emph{First}, they enable assessment of whether a VQA model reasons based on the correct visual evidence.  This is valuable as an explanation as well as to support developers in debugging models.  \emph{Second}, answer groundings enable segmenting the relevant content from the background.  This is a valuable precursor for obfuscating the background to preserve privacy, given that photographers can inadvertently capture private information in the background of their images~\cite{gurari2019vizwiz} (exemplified in Figure~\ref{fig:use-cases}b).  \emph{Third}, users with low vision could more quickly find the desired information if a service instead magnified the relevant visual evidence.  This is valuable in part because answers from VQA services can be insufficient, including because humans suffer from ``reporting bias" meaning they describe what they find interesting without understanding what a person/population is seeking.  This is exemplified in Figure~\ref{fig:use-cases}c, where the most popular response from 10 answers is the generic answer `pasta' rather than the specific flavor, `Creamy Tomato Basil Penne'.  

While datasets have been introduced to encourage progress on the answer grounding problem, all proposed dataset challenges originate from contrived visual questions~\cite{Zhu_2016_CVPR,das2017human,gan2017vqs,krishna2017visual,huk2018multimodal,hudson2019gqa,chen2020air,urooj2021found,nagaraj-rao-etal-2021-first}.  This includes scraping images from photo-sharing websites (e.g., Flickr) and then generating questions automatically~\cite{chen2020air}, by using image annotations paired with question templates to create questions about the images or (2) manually~\cite{das2017human,gan2017vqs,Zhu_2016_CVPR,huk2018multimodal}, by asking crowdworkers to make up questions about an image that would stump a robot.  Yet, prior work has shown that such contrived settings can manifest different characteristics from authentic VQA use cases~\cite{gurari2018vizwiz,zengvision}.  This can cause algorithms trained and evaluated on contrived datasets to perform poorly when deployed for authentic use cases~\cite{gurari2018vizwiz,gurari2020captioning}.  Moreover, this can limit the designs of algorithms since developers are oblivious to the additional challenges their algorithms must overcome.  

We introduce the first answer grounding dataset that originates from an authentic use case.  We focus on visual questions originating from blind people who both took the pictures and asked the questions about them in order to overcome real visual challenges~\cite{bigham2010vizwiz}.  This use case has been shown to manifest different challenges than contrived settings, including that images are lower quality \cite{chiu2020assessing}, questions are more conversational~\cite{gurari2018vizwiz}, and different vision skills are needed to arrive at answers~\cite{zengvision}. For approximately 10,000 image-question pairs submitted by this population, we collected answer groundings. Then, we analyzed the answer groundings to reveal their characteristics and show how they relate/differ to five existing answer grounding datasets. Finally, we benchmarked state-of-the-art VQA and answer grounding models on our dataset and demonstrate what makes this dataset difficult for them, including smaller answer groundings, images that are of higher quality, and visual questions that require skills in text recognition. 

We offer this work as a foundation for designing models that are robust to a larger range of potential challenges that can arise in real-world VQA settings.  Challenges observed in our dataset can generalize to other scenarios, such as robotics and lifelogging, which similarly encounter varying image quality and textual information (e.g., grocery stores).  To encourage community-wide progress on such challenges, we have organized a dataset challenge with public evaluation server and leaderboard. Details can be found at the following link: \url{https://vizwiz.org/tasks-and-datasets/answer-grounding-for-vqa/}. 

\section{Related Work}

\paragraph{VQA Datasets.}
Many large-scale VQA datasets have been proposed over the past six years~\cite{wu2017visual,Kafle_2017_ICCV,srivastava2019visual,srivastava2020visual}.  A key challenge the community has faced in developing such datasets is the \textit{language bias} problem \cite{goyal2017making, niu2021counterfactual,kv2020reducing, ramakrishnan2018overcoming}.  In particular, models can learn to exploit superficial correlations observed in datasets by identifying common pairings of answers and questions and never looking at the images. For example, when given an image of a green banana with an associated question asking “what is the color of the banana?”, the VQA model might answer “yellow” instead of green because the answer ``yellow" appears more frequently in the answers for questions related to bananas.  This problem has been possible, in part, because common evaluation metrics \cite{antol2015vqa} only assess model performance based on textual answers.  To combat such biases, new VQA datasets have been created to more equally capture the range of possible answers for each question~\cite{balanced_vqa_v2} and many new models have been developed using these balanced datasets.  Our work contributes to this body of work by enabling algorithm developers to go beyond only evaluating the textual answers and also directly evaluate whether algorithms rely on the correct visual evidence.

\vspace{-0.75em}\paragraph{Answer Grounding Datasets. }
Towards disentangling the vision problem for VQA, several answer grounding datasets already have been introduced that locate the visual content needed to arrive at each language-based answer~\cite{das2017human,gan2017vqs,Zhu_2016_CVPR,huk2018multimodal,chen2020air}.  While some were created by tracking where humans look when presented with a visual question~\cite{das2017human,chen2020air} or collecting bounding boxes around the relevant visual evidence~\cite{Zhu_2016_CVPR, krishna2017visual,hudson2019gqa}, our work more closely aligns with those where humans who annotated the relevant visual evidence using a segmentation~\cite{gan2017vqs, huk2018multimodal,nagaraj-rao-etal-2021-first}.  That is because we also collect segmentations.  We propose the first answer grounding dataset that reflects an authentic VQA use case and conduct extensive analysis to demonstrate how it relates/differs to five existing answer grounding datasets.

\vspace{-0.75em}\paragraph{VQA and Answering Grounding Algorithms.} 
Modern VQA algorithms often rely on attention maps to determine where to look to find the answers to visual questions, with this having been the dominant approach for a number of years~\cite{das2017human, qiao2018exploring,zhang2019interpretable}.  Recently, some researchers have shifted their focus towards delivering the best VQA models possible under the constraint that the model should outperform all other VQA models in attending to the correct visual evidence \cite{urooj2021found}.  We benchmark the state-of-the-art VQA and answer grounding models on our dataset to examine to what extent they succeed in correctly grounding answers.  Experiments reveal that our new dataset is challenging for modern algorithms, and show what aspects make it challenging. 


\vspace{-0.75em}\paragraph{Assistive Technology for People with Vision Impairments.}
Many people with visual impairments rely on visual assistance devices to learn about their surroundings.  For instance, people with low vision often rely on magnification tools to better observe content of interest~\cite{kline1995improving,polzer2018gaze,schwarz2020developing}, given that they have limited sight that can't be regained fully with corrective measures such as glasses.  In addition, people with low vision and no vision rely on on-demand technologies~\cite{bigham2010vizwiz,bemyeyes,bespecular} that deliver answers to submitted visual questions.  A challenge for this latter use case is that blind people have no way to check if they inadvertently capture private information in their images.  Yet, roughly 12\% of pictures taken for a VQA use case contained privacy information~\cite{gurari2019vizwiz} and visually impaired people have expressed their discomfort with leaking private information~\cite{akter2020uncomfortable,stangl2020visual,stangl2022privacy}.  Our work can contribute to a wide range of interests for people with vision impairments since answer groundings can serve as a valuable precursor to smartly magnify visual answers to visual questions, mitigate biases in VQA services, and support increased privacy measures with obfuscation.
    
\section{VizWiz-VQA-Grounding Dataset}
We now introduce our dataset for grounding answers to visual questions asked in an authentic use case where people who are blind were trying to learn about their visual surroundings. We call this dataset ``VizWiz-VQA-Grounding".

\subsection{Dataset Creation}
\label{section:datasetcreation}
\paragraph{Dataset Source.}
Our work builds upon the VizWiz-VQA dataset~\cite{gurari2018vizwiz}, which consists of 32,842 image question pairs where each comes with 10 crowdsourced answers. In addition to the VQA triplets for the publicly-available train and validation splits, we also consider those from the test split since the authors of~\cite{gurari2018vizwiz} provided us with answer annotations for this purpose. The images and questions come from visually impaired people who shared them to solicit visual assistance in their daily lives.  


\vspace{-0.75em}\paragraph{Dataset Filtering.}
We designed our dataset to focus on grounding answers for visual questions that could unambiguously be grounded to a single region. Towards achieving this, we filtered the initial dataset using a combination of automated and manual techniques that are described in the Supplementary Materials.  In summary, we removed all questions that were non-answerable, embedded multiple sub-questions, referred to multiple regions in an image due to ambiguity, could not be grounded, or for which the majority of crowd did not agree on a single answer. This process left a total of 9,998 VQAs, which we use for our new dataset. We focus on grounding only the single most popular answer for each visual question.

\vspace{-0.75em}\paragraph{Grounding Task Design.}
Through iterative pilot studies, we designed a user interface for grounding answers to visual questions. A person is shown the question-answer pair just above the image and must then demarcate the answer grounding by clicking a series of points on the image to create a connected polygon.  We provide extensive instructions to cover many annotation scenarios that may be tricky. For instance, when multiple regions containing the same thing need to be annotated (e.g., a bunch of flowers), we instruct the annotator to demarcate all relevant content with a single polygon when the regions are connected.  When the question asks about a property of a visual entity (e.g., color of clothes), we instruct the annotator to comprehensively annotate all regions that would lead to the answer rather than a minimum viable region.

\vspace{-0.75em}\paragraph{Annotation Collection.}
We implemented a number of techniques to support collecting high quality results.  First, we developed a rigorous three-step filtering process to recruit expert crowdworkers from Amazon Mechanical Turk to complete our annotation tasks, which we describe in the Supplementary Materials.  In summary, we limit which users can complete our tasks, provide a qualification test, and then select several expert workers who we identified as excelling at our task for a larger number of tasks.  Next, for all work submitted by these expert workers, we conduct automated and manual quality control to verify that we can continue to trust their work.  Finally, we collected two answer groundings per VQA instance (i.e. image-question-answer triplet).  Based on our subsequent analysis for when and why answer groundings differed, we decided to select the larger of the two annotations as our ground truth annotation because we found that annotation differences typically occurred because one was a sub-region of the other.

\subsection{Dataset Analysis}
\label{section:dataAnalysis}
We now analyze the VizWiz-Visual Grounding dataset, which consists of 9,998 groundings for the 9,998 VQA triplets. To do so, we compute for each grounding its: 
\begin{itemize}
    \setlength\itemsep{0.1em}
    \item \textbf{Location}: position of its center of mass relative to the entire image; i.e., a (x,y) coordinate.  Each coordinate can range from 0 to 1. 
    \item \textbf{Boundary complexity}: entropy of the histogram of the normalized \textit{centroid contour distance}~\cite{chen2005estimating}, where \textit{centroid contour distance} is the distance of every point on the segmentation boundary to the segmentation's center of mass.  Values can range from 0 to 1. 
    \item \textbf{Image coverage}: fraction of pixels it occupies from all pixels in the image.  Values can range from 0 to 1.
\end{itemize}

To compare our dataset to the current focus of the research community, we also evaluate answer groundings from existing datasets that were similarly generated through a manual segmentation annotation process: VQS \cite{gan2017vqs}, VQA-X \cite{huk2018multimodal}, and TextVQA-X \cite{nagaraj-rao-etal-2021-first}.  For completeness, we also include CLEVR-Answers \cite{urooj2021found} and GQA \cite{hudson2019gqa}, as these groundings were used to validate the state-of-the-art answer grounding method~\cite{urooj2021found}.  Unlike our dataset, the images for all these datasets are either scraped from the Internet---including VQS, VQA-X, and GQA which leverage images from COCO~\cite{lin2014microsoft} and TextVQA-X which leverages images from Open Images v3 \cite{openimages}---or computer generated, as is the case for CLEVR-Answers.  Another difference from our dataset is that the questions for these datasets were either generated by crowd workers ---for VQS, VQA-X, TextVQA-X---or computer generated, as is the case for GQA and CLEVR-Answers.  To support fair comparison, we only consider visual questions in these datasets for which there is exactly one answer grounding region.  

\begin{figure}[t!]
    \hspace*{-5mm}
    \includegraphics
    [scale=0.261]{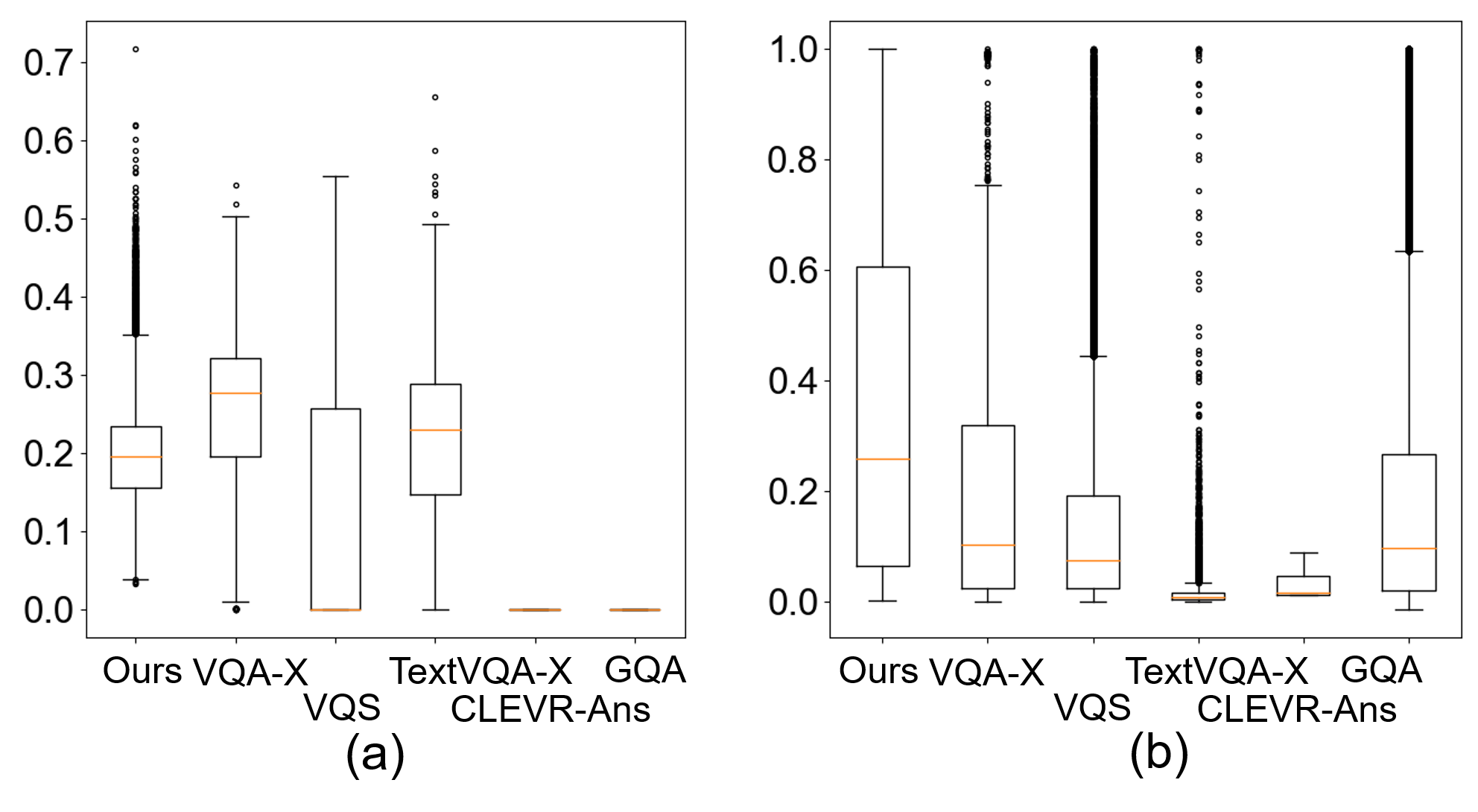}
    \vspace{-2em}
    \vspace{-0.25em}\caption{The box plot shows, for each dataset, the range of values for the answer groundings' (a) boundary complexity and (b) image coverage.  For each box, the central mark denotes the median score, the box edges denote the 25th and 75th percentiles scores, the whiskers denote the most extreme data points not considered outliers, and the outliers are plotted individually.  Overall, our dataset shows a moderate range for boundary complexity and the greatest range for answer grounding size.}
    \label{figure:dataset-statistics}
\end{figure}

\vspace{-0.75em}\paragraph{Overall Results.}
For all datasets, statistics about the answer groundings' \emph{location} are shown in Table~\ref{table:center-of-mass} and \emph{boundary complexity} and \emph{image coverage} are shown in Figure~\ref{figure:dataset-statistics}. 

Overall, we observe that all datasets have answer groundings that typically lie close to the center of the image (Table~\ref{table:center-of-mass}).  This is evident from the combination of mean centroids around the coordinates (0.5, 0.5) and relatively small standard deviations from those coordinates.  We found this result surprising for our dataset since visually impaired photographers cannot verify they center the content of interest when taking the pictures.  

Regarding the complexity of the boundaries for answer groundings, our dataset lies in the middle of all datasets.  On one extreme lies both datasets for which bounding boxes are collected: CLEVR-Ans and GQA.  That is because the value computed for boundary complexity is 0 when the boundary is a rectangle.  Interestingly, we observe for VQS that over half of the answer groundings also are rectangles, as shown by its median score of 0.  At the other extreme, lies the following datasets: VQA-X and TextVQA-X.  We suspect the higher boundary complexity for TextVQA-X is due to its annotation collection approach.  Specifically, it is the only dataset which collected groundings using a paintbrush rather than a series of points clicked around the boundary of an object.  Implicitly, this annotation approach leads to a seemingly more complex boundary when a simple polygon might suffice.  Our dataset is most similar to VQA-X, which has the highest median complexity, but our dataset also exhibits a smaller range of complexity values.  

\begin{table}[t!]

\centering
\begin{tabular}{lc}
\hline
\textbf{}   & \textbf{Relative location of groundings} \\ \hline
\textbf{Ours} & (0.48$\pm$0.14, 0.51$\pm$0.15)                                        \\ 
\textbf{VQA-X}       & (0.50$\pm$0.16, 0.53$\pm$0.19 )                                \\ 
\textbf{VQS}         & (0.50$\pm$0.20, 0.52$\pm$0.21)                                   \\ 

\textbf{TextVQA-X}    & (0.49$\pm$0.21 , 0.48$\pm$0.25)                              \\ 
\textbf{CLEVR-Ans} & (0.50$\pm$0.19, 0.45$\pm$0.13)  \\ 
\textbf{GQA} & (0.50$\pm$0.20, 0.54$\pm$0.19)   \\ \hline
\end{tabular}
\vspace{-0.5em}
\caption{Shown is the mean and standard deviation of the location of all answer groundings with respect to the images for each dataset.  Across all datasets, answer groundings tend to be located near the center of the image.}
\label{table:center-of-mass}
\end{table}

\begin{figure}[t!]
    \centering
    \includegraphics [scale=0.6]{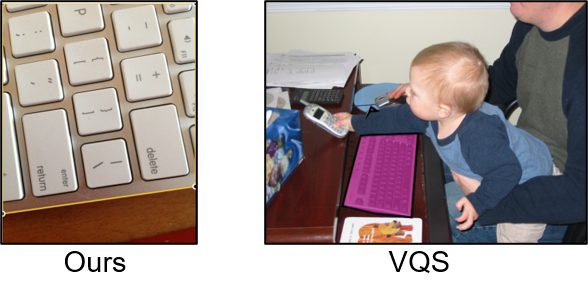}
    \vspace{-1em}\caption{Answer grounding for a ``keyboard" in our dataset and in the VQS dataset. This exemplifies that answer groundings in our dataset can pertain to large regions in the images since photographers in an authentic use case zoom into the content of interest to try and only photograph the pertinent content.}
    \label{fig:keyboard}
\end{figure}

The key criteria that makes our dataset different from the other datasets, overall, is that its answer groundings occupy a larger range of image coverage values than all other datasets.  This is evident when comparing the box sizes for all datasets in Figure ~\ref{figure:dataset-statistics}(b).  TextVQA-X occupies the smallest region. We suspect that VQA-X, VQS, and GQA share similar sizes because they all were generated from the same image source: COCO images.  We attribute our dataset's inclusion of many considerably larger answer groundings than the other datasets to the possibility that \textit{zooming into the content of interest} when taking the picture may be a more realistic approach photographers would take in real-world scenarios as they try to only photograph the pertinent content for answering the question.  This distinction is exemplified in Figure \ref{fig:keyboard}, where the visually impaired photographer took a photo by positioning a keyboard close to the hand which held the camera.  Altogether, this finding underscores a unique benefit of our dataset in that it motivates the design of algorithms that can simultaneously locate very large and very small regions.

\begin{table*}[t!]\scalebox{0.8}{
\begin{tabular}{lccccc}
\hline
\textbf{}   & \textbf{What is this} & \textbf{What color is this} & \textbf{What color is this shirt} & \textbf{What is in this box} & \textbf{What does this say} \\ \hline
\textbf{Ours} &   (0.48$\pm$0.11,0.51$\pm$0.11)       &  (0.50$\pm$0.07,0.51$\pm$0.08)    &(0.50$\pm$0.06,0.50$\pm$0.07)    &    (0.55$\pm$0.12,0.64$\pm$0.15)     &        (0.48$\pm$0.14,0.47$\pm$0.16)         \\ 
\textbf{VQS}         &    (0.69$\pm$0.09,0.45$\pm$0.09)     & (0.8,0.46)   & - & -      &  -        \\ 

\textbf{TextVQA-X}    & (0.47$\pm$0.37,0.56$\pm$0.13)                & - & -  & -  &  (0.62,0.21)     \\ 
\hline 
\end{tabular}}
\vspace{-0.75em}\caption{Shown is the mean and standard deviation of the location of all answer groundings with respect to the images for each dataset for the five most common questions observed in our dataset.  ``-" denotes that no statistics were computed because the dataset lacks the question.  The results highlight that different questions often have different typical locations, especially across different datasets.}
\label{table:center-of-mass-MostCommonQuestion}

\end{table*}

\begin{figure}[t!]
    \includegraphics [width= 0.95\linewidth]{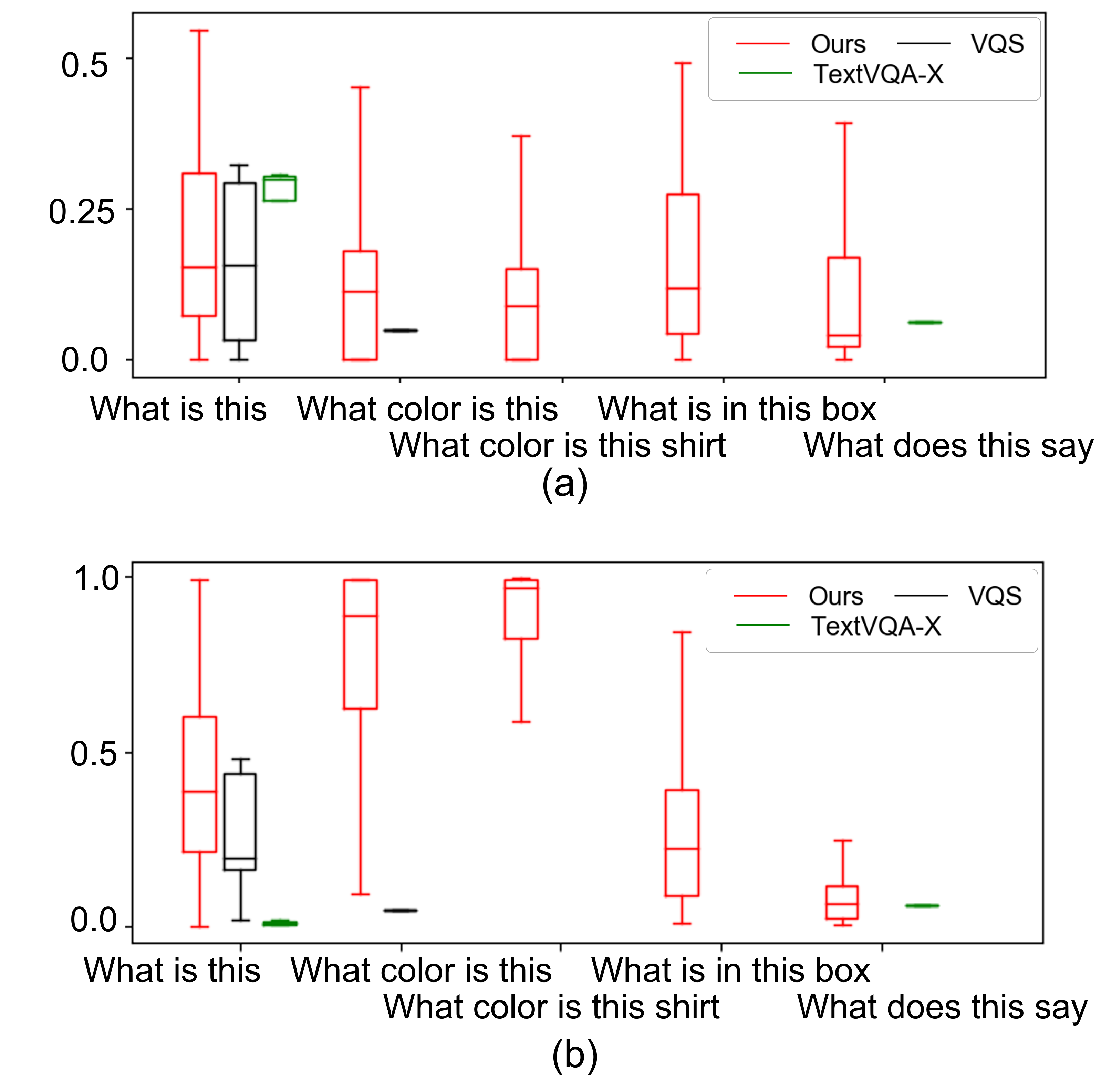}
    \vspace{-1em}\caption{For each dataset, the box plot shows the range of values for the answer groundings for the most common \emph{questions} in our dataset with respect to: (a) boundary complexity and (b) image coverage.  (Figure~\ref{figure:dataset-statistics} describes the box plot visualization)}  
    \label{fig:MostCommonQuestion}
\end{figure}

\vspace{-0.75em}\paragraph{Most Common Questions.}
We next evaluate how our dataset compares with existing answer grounding datasets for the \textbf{five most common visual questions} from our VizWiz-VQA-Grounding dataset. The most common questions are: ``What is this", ``What color is this", ``What color is this shirt", ``What is in this box", and ``What does this say".\footnote{We group the following questions together: ``what is this", ``what is it", ``what is this item", ``what is that", ``what's that", ``what is this please" into the ``What is this" group. We also group the following questions together: ``what color is this" and ``what color is it". } Note that none of these questions are observed in the VQA-X, CLEVR-Answers, and GQA datasets. Consequently, we exclude these three datasets from comparison.  For each dataset, analysis of the groundings with respect to these questions are shown in Table~\ref{table:center-of-mass-MostCommonQuestion} for \emph{location} and Figure~\ref{fig:MostCommonQuestion} for \emph{boundary complexity} and \emph{image coverage}. 

Across all datasets, we observe different types of questions manifest different typical locations, boundary complexities, and sizes from each other.  For instance, in Figure~\ref{fig:MostCommonQuestion}(b), we see that the average boundary complexity for ``What does this say" is uniquely near zero for our dataset and so usually can be grounded with a rectangle.  We suspect algorithms will be able to take advantage of these differences to learn predictive cues for grounding answers. 

We also observe that the characteristics of answer groundings for the same question are considerably different across the different datasets.  This is evident when examining the results for ``What color is this" and ``What does this say"; i.e., there are considerable differences for the typical location of the answer grounding (Table~\ref{table:center-of-mass-MostCommonQuestion}), the typical range of boundary complexity values (Figure~\ref{fig:MostCommonQuestion}a), and the typical range of image coverage values (Figure~\ref{fig:MostCommonQuestion}b).  Consequently, if models trained on other datasets are learning biases between specific questions and answer grounding locations without truly understanding the question, they would generalize poorly to our new dataset (and vice versa).

\begin{figure}[t!]
    \includegraphics
    [width=0.92\linewidth]{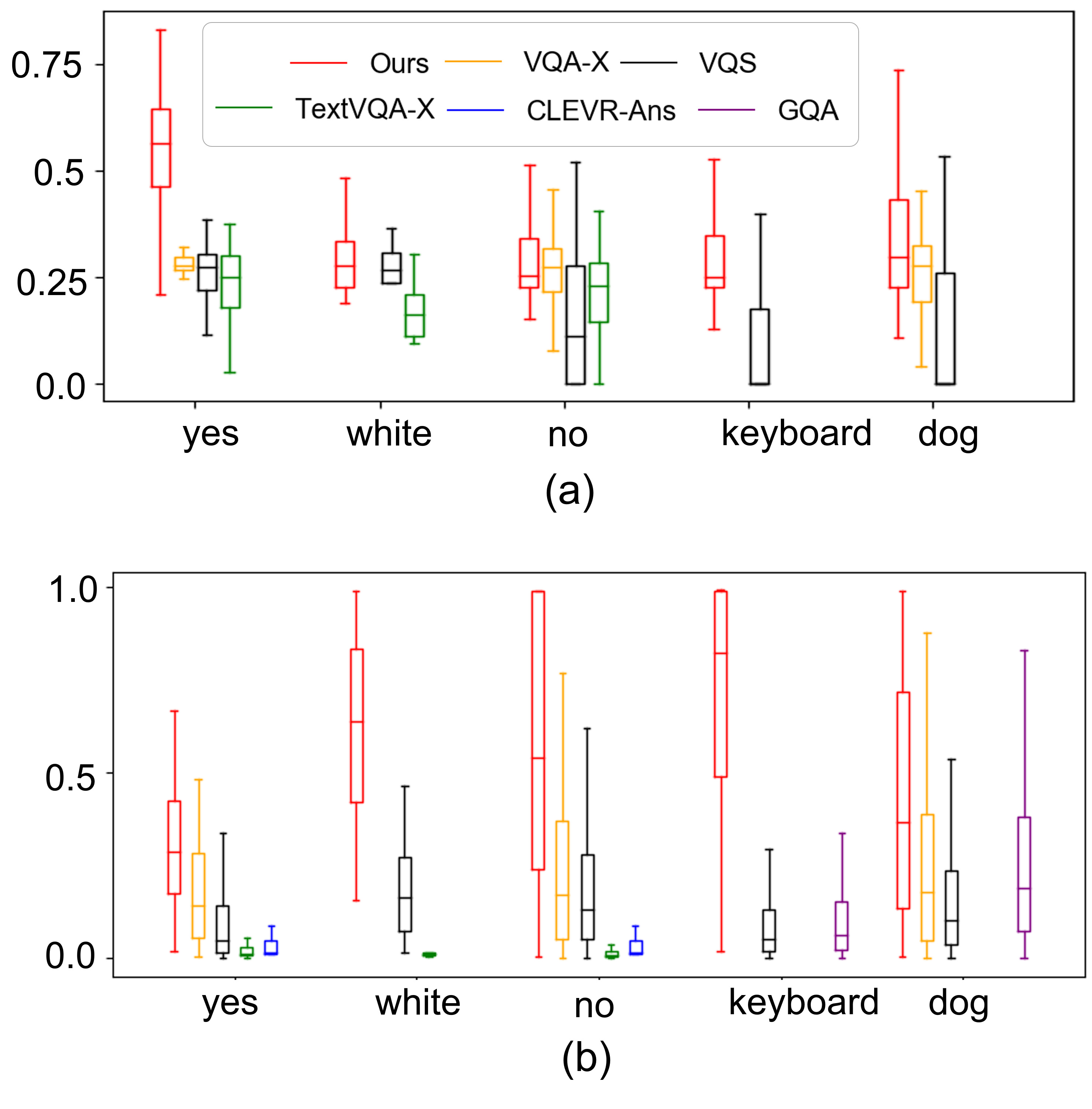}
    \vspace{-1em}\caption{For each dataset, the box plot shows the range of values for the answer groundings for the most common \emph{answers} in our dataset with respect to: (a) boundary complexity and (b) image coverage.  (Figure~\ref{figure:dataset-statistics} describes the box plot visualization)}
    \label{fig:MostComonAnswer}
\end{figure}

\vspace{-0.75em}\paragraph{Most Common Answers.}
We perform parallel analysis to what we conducted for the most common questions but now with respect to \textbf{five of the most common answers} from the VizWiz-VQA-Grounding dataset: ``yes", ``white", ``no", ``keyboard", and ``dog".\footnote{We restricted our analysis to one color-related answer to support greater diversity in our analysis.} For each dataset, we analyze groundings with respect to these specific answers.  
We observe that the relative location is similar for all answers, with answers typically grounded at the center of the images.  Due to space constraints, we include these results in the Supplementary Materials.  In contrast, we observe that different answers manifest different statistics for the boundary complexity (Figure~\ref{fig:MostComonAnswer}a) and relative sizes (Figure~\ref{fig:MostComonAnswer}b), both within each dataset and across different datasets.  This reinforces our findings from the most common questions.  Answer differences can yield valuable predictive cues for grounding answers and cross-dataset evaluation could be valuable for preventing models from learning superficial correlations between answers and answer grounding characteristics in a specific dataset.

\begin{figure}
    \includegraphics [width=\linewidth]{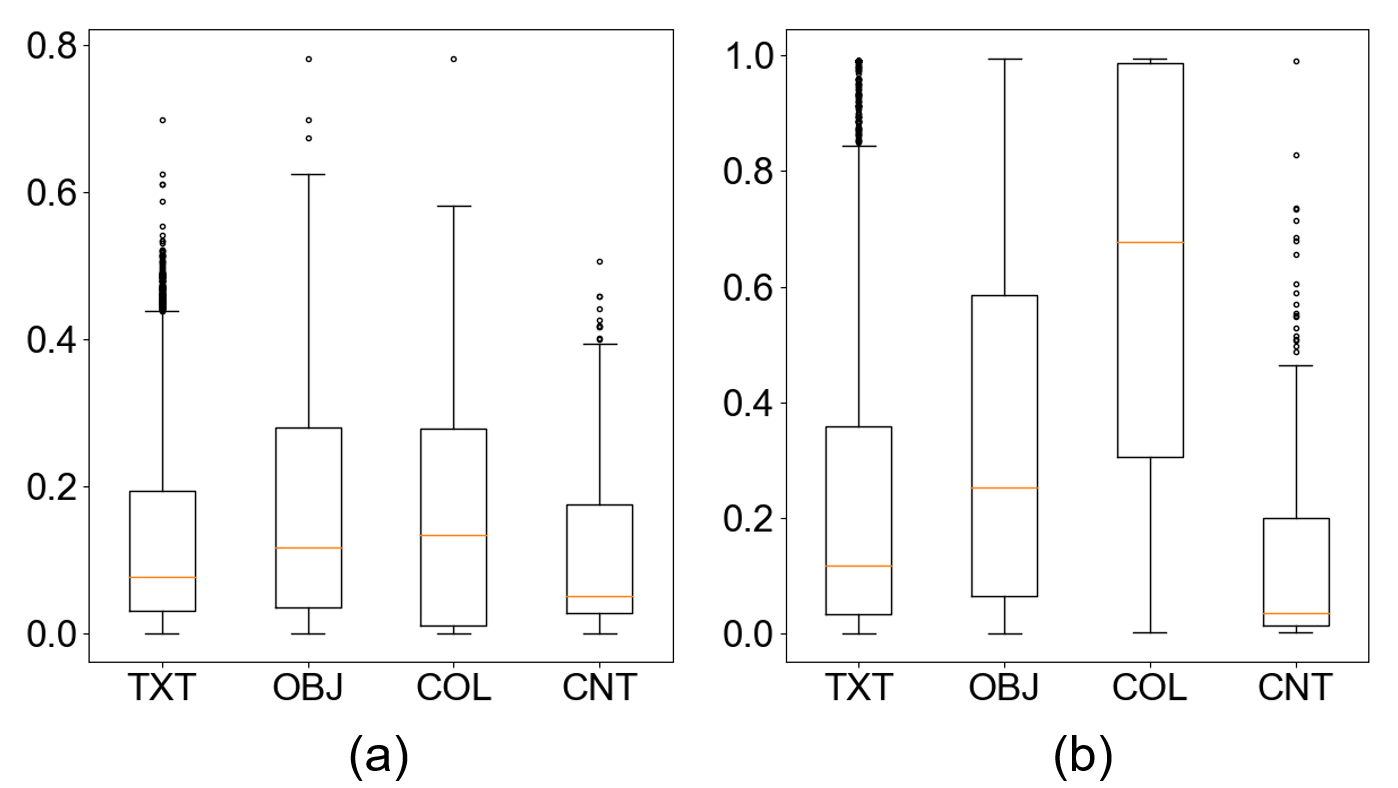}
    \vspace{-1.5em}\caption{The box plot shows the range of values for (a) boundary complexity and (b) image coverage for all answer groundings in our dataset with respect to visual questions that require different vision skills. (Figure~\ref{figure:dataset-statistics} describes the box plot visualization)}
    \label{fig:visionskills}
\end{figure}

\vspace{-0.75em}\paragraph{Vision Skills Needed to Answer Visual Questions.}
\label{section:visionskill}
We also conduct fine-grained analysis of our dataset with respect to the types of skills needed to answer visual questions, using the labels provided in the VizWiz-VQA-Skills dataset~\cite{zengvision}.  Overall, we again observe that answer groundings typically are positioned at the center of the image with respect to vision skills. Due to space constraints, we include these results in the Supplementary Materials.  In contrast, 
we observe that different vision skills are correlated with different statistics for image coverage (Figure~\ref{fig:visionskills}b).  Consequently, vision skills also could provide valuable predictive cues that algorithms could latch onto when trying to ground answers.  For instance, visual questions about trying to read \textit{text} tend to have a relatively small visual grounding area in the image and, often, can be grounded with a simple bounding region (e.g., with four points).  In contrast, questions related to recognizing \textit{color} tend to have a larger visual grounding area and more complex boundary.  We attribute this distinction to text being a consistently well-structured entity with clear printing boundaries whereas color can be used to describe less structured entities such as articles of clothing which can manifest a variety of shapes.

\section{Automatically Grounding Answers for VQA}
\label{section:algorithm}
Using our grounding dataset, we now quantify to what extent state-of-the-art VQA models look at the correct regions in images to answer the questions. 

\vspace{-0.75em}\paragraph{Dataset Splits.} 
Our VizWiz-VQA-Grounding dataset's train/val/test/ splits match the train/val/test splits of the original VizWiz-VQA dataset~\cite{gurari2018vizwiz}.  This resulted in 6,494, 1,131 and 2,373 visual questions in the training, validation, and test sets respectively.    

\vspace{-0.75em}\paragraph{Baseline Models.} 
We benchmark a total of six models.  

First, we chose the top-performing VQA algorithms with publicly-available code for the 2021 VizWiz-VQA dataset challenge~\cite{gurari2018vizwiz} and the mainstream 2021 VQA dataset challenge~\cite{balanced_vqa_v2}, that is   \emph{LXMERT} \cite{tan2019lxmert} and \emph{OSCAR} \cite{li2020oscar} \emph{with VinVL image features}~\cite{zhang2021vinvl} respectively.  Both models are pretrained on the train splits of their respective challenge datasets.  In order to generate attention masks, we follow the process described in \cite{tan2019lxmert} to analyze attention maps extracted from each model. Using the default parameters, attention weights across the multiple attention heads are extracted and averaged to obtain the final attention map. A threshold of 0.5 is then applied to generate the final binary segmentation mask.   

We also chose the state-of-the-art model for answer grounding: MAC-Caps \cite{urooj2021found}.  Given an image and question, MAC-Caps predicts an answer and attention weights on the image.  As done in \cite{urooj2021found}, we obtain the final binary segmentation mask by applying a threshold of 0.5 to the attention weights extracted from the last reasoning step.  We benchmark four variants of MAC-Caps. We use the two models that were pretrained on GQA and CLEVR respectively, as described in the original paper. Next, we train the MAC-Caps algorithm from scratch using the train split of the VizWiz-VQA \cite{gurari2018vizwiz}  dataset.  Finally, we also train the MAC-Caps algorithm from scratch using the train split of the VQA-v2 dataset \cite{balanced_vqa_v2}, since this dataset was by design intended to prevent models from learning language biases and to instead encourage models to look at the images.   

\vspace{-0.75em}\paragraph{Evaluation Metric.} We employ Intersection over Union (IoU) to measure the similarity of each binary segmentation mask  to the ground truth segmentation. Values range from 0 to 1, with higher values indicating better performance.  We compute the mean IoU score across all test examples and report results are percentages (i.e., IoU value x 100).

We also evaluate with the common metric for detection and localization tasks: mAP@IoU.  Following the COCO evaluation protocol, we use different IoU thresholds, from 0.25 to 0.75, and average AP values with IoU thresholds in the range of 0.5 to 0.95 with a step size of 0.05.  Due to space constraints, results are provided in the Supplementary Materials.  In summary, these results reinforce our findings with respect to the IoU metric, as described below.

\vspace{-0.75em}\paragraph{Overall Results.}
The performance for each model on the VizWiz-VQA-Grounding test split is shown in Table \ref{table:MAC-Caps results}. Performance is reported for all visual questions (column 3) as well as only the subset of visual questions for which each model correctly predicted the answers (column 4).

\begin{figure*}[t!]
     \centering
     \begin{subfigure}[b]{0.27\textwidth}
         \centering
         \includegraphics[width=\textwidth]{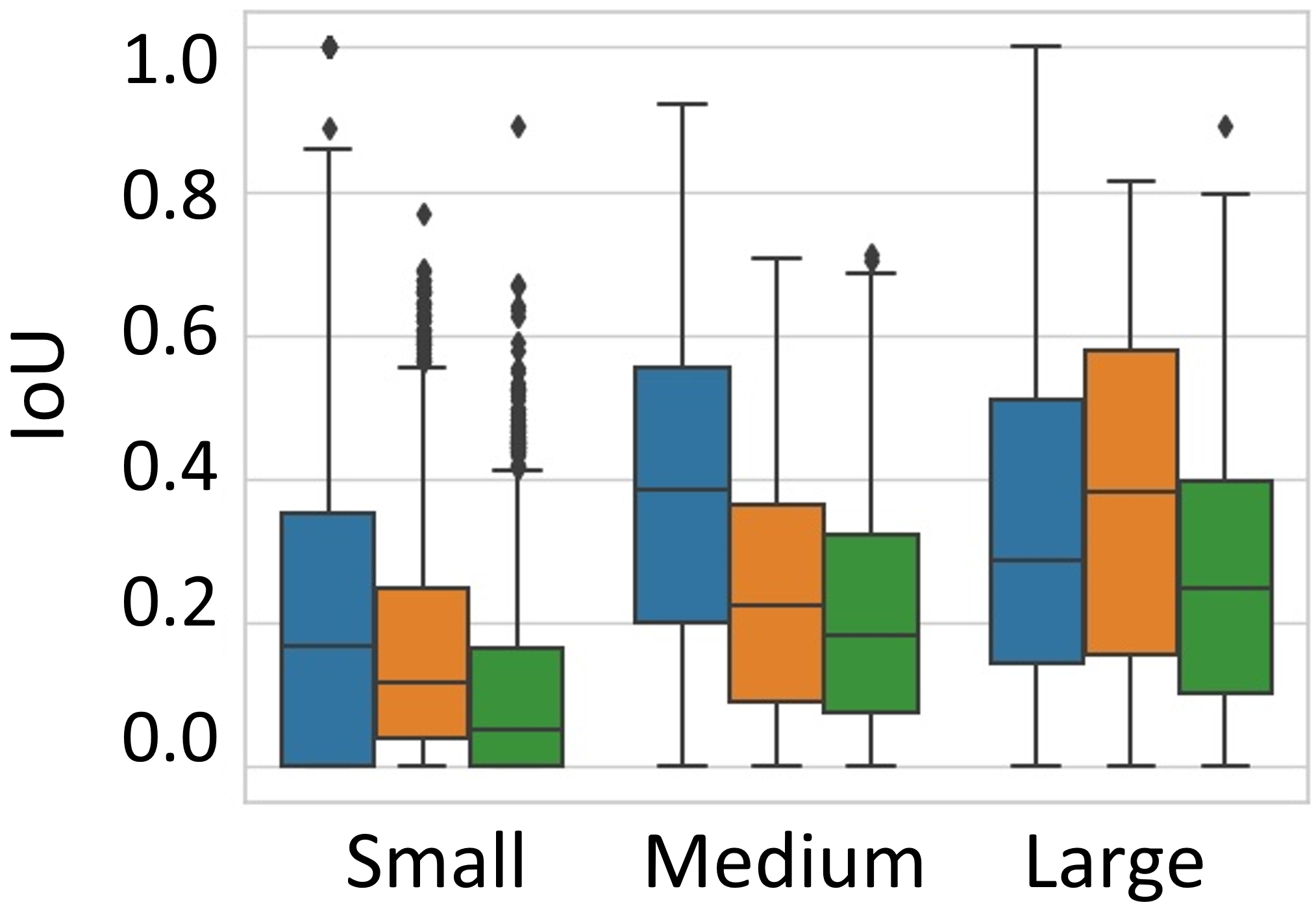}
         \caption{Image Coverage}
       \label{fig:AlgorithmImageCoverage}
     \end{subfigure}
     \hfill
     \begin{subfigure}[b]{0.27\textwidth}
         \centering
         \includegraphics[width=\textwidth]{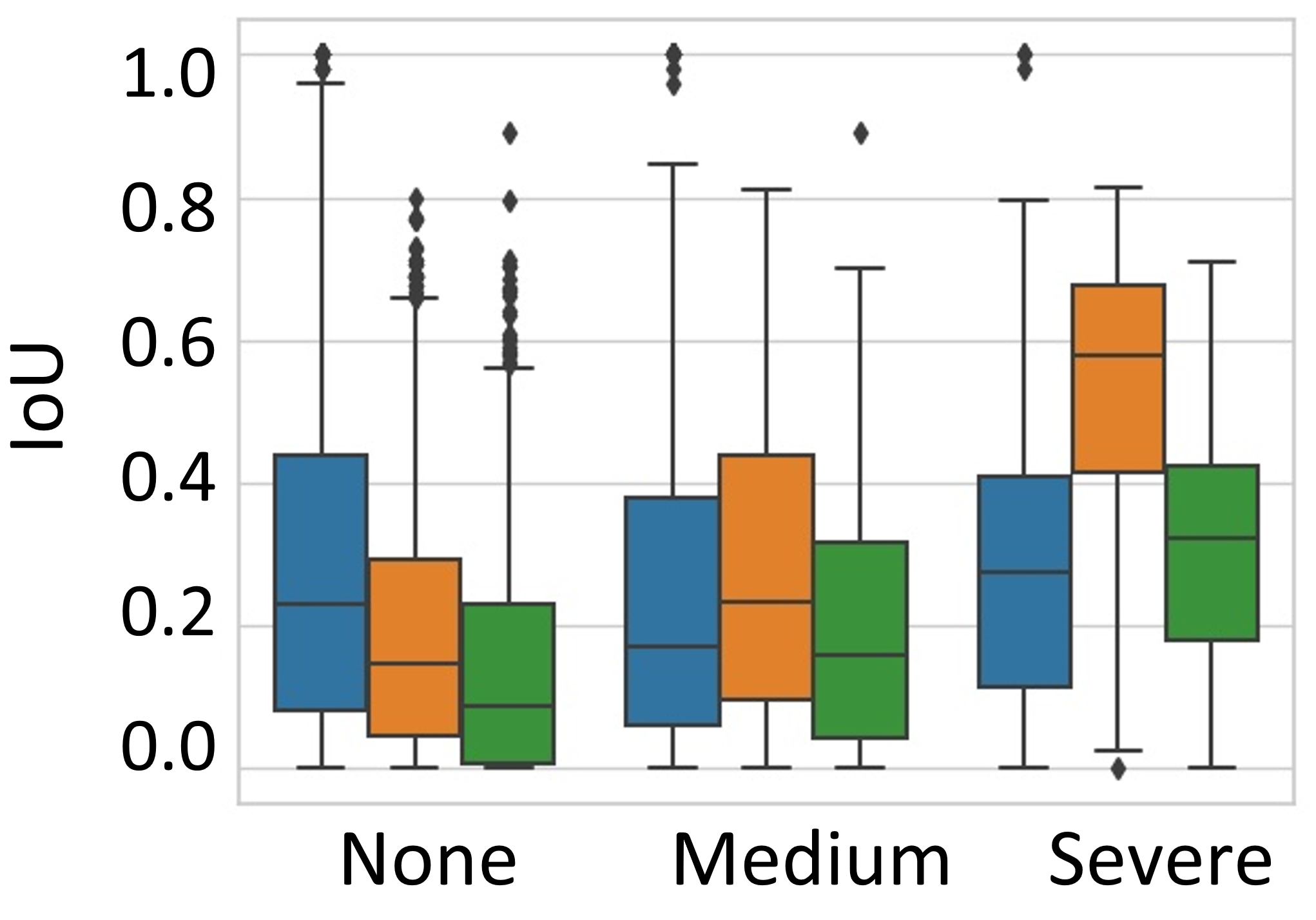}
         \caption{Image Quality}
         \label{fig:AlgorithmImageQuality}
     \end{subfigure}
     \hfill
     \begin{subfigure}[b]{0.40\textwidth}
         \centering
         \includegraphics[width=\textwidth]{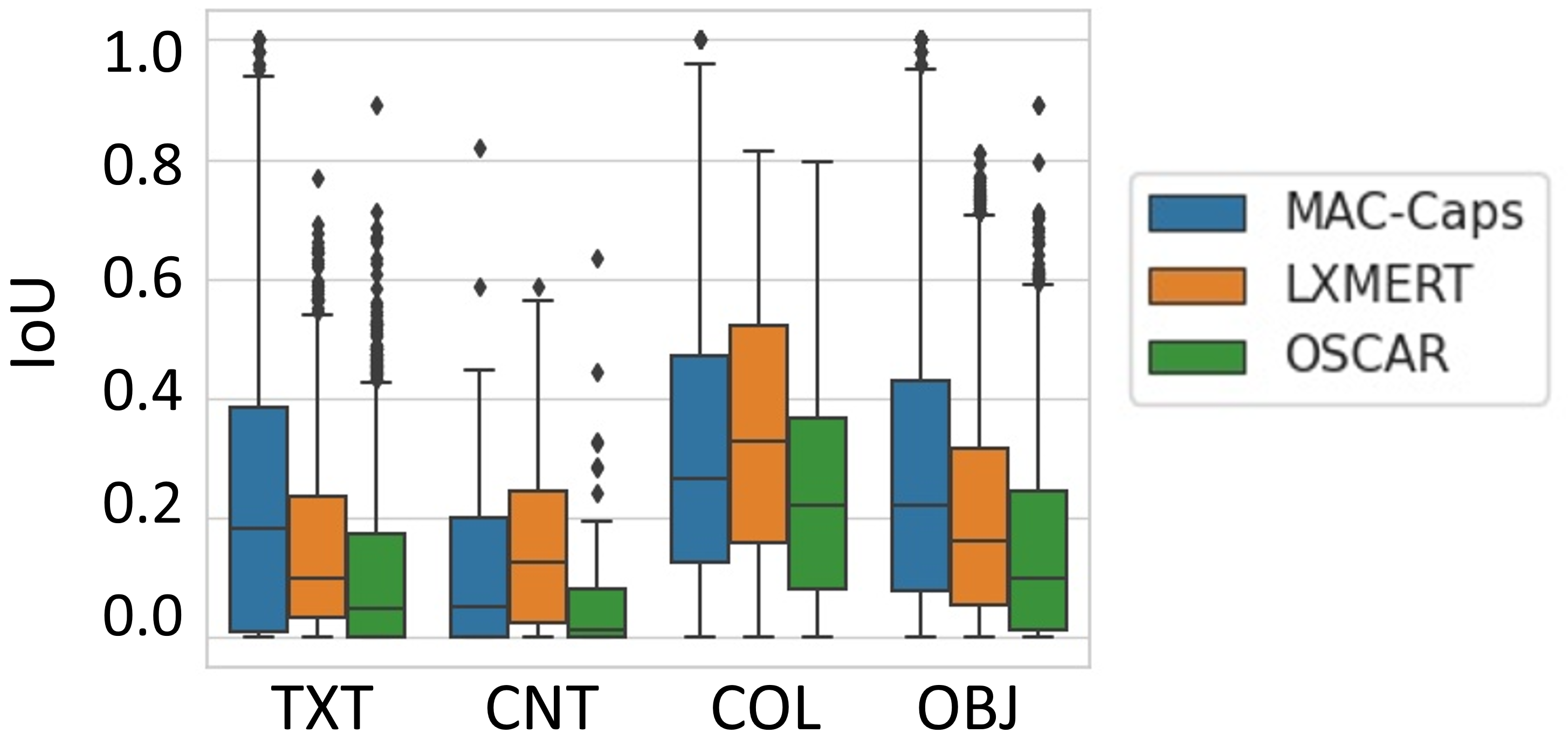}
         \caption{Vision Skills}
         \label{fig:AlgorithmSkill}
     \end{subfigure}
     \hfill

      \vspace{-0.25em}\caption{Comparison  of  MAC-Caps  (pretrained  on VizWiz),  LXMERT,  and  OSCAR’s  performance  on (a) visual questions containing different sizes of attention areas for the answers, (b) visual questions for images with different severity of quality issues, and (c)  visual questions  requiring  different  vision  skills  to  answer  correctly.  Overall, we observe (a) smaller answer groundings, (b) images with no quality issues, and (c) visual questions requiring counting skills are most challenging for the models. }
        \label{fig:analysis }
\end{figure*}

\begin{table}[t!]
\small
\begin{tabular}{cccc}
\hline
\textbf{Model} & \textbf{Pretrained}   & \textbf{Avg IoU} & \textbf{Avg IoU (correct)}\\\hline
LXMERT & VizWiz-VQA       &           22.09              &    26.96 (\textbf{906})     \\ 
OSCAR & VQA-v2       &   15.48 &    19.79 (693)   \\ 
MAC-Caps & GQA &      12.56         &    17.77 (270)         \\     
MAC-Caps & CLEVR        &        15.31                  &  10.98 (60)           \\ 
MAC-Caps & VQA-v2       &           17.42              &    19.58 (374)    \\ 
MAC-Caps & VizWiz-VQA    &            \textbf{27.43}                &    \textbf{32.8} ({352})   \\   \hline
\end{tabular}
\vspace{-0.25em}\caption{Performance of six models when evaluated on the VizWiz-VQA-Grounding test set: two state-of-art VQA models (LXMERT~\cite{tan2019lxmert} and OSCAR~\cite{li2020oscar}) and four variants of the state-of-art VQA model for answer grounding (MAC-Caps~\cite{urooj2021found}). IoU scores (averaged over all 2,373 samples for each model) are reported in percentages. The number in parentheses is the total number of answers correctly predicted by each model.}
\label{table:MAC-Caps results}
\end{table}

We observe that all models performed poorly overall.  For example, the top-performing MAC-Caps model that was trained on the VizWiz-VQA dataset achieves an IoU score of 27.43\%.  The story improves only modestly when considering just those visual questions for which the model predicted correct answers; i.e., the IoU jumps $\sim$5 percentage points to 32.8\%. These findings indicate that existing mechanisms intended to guide models to look at the correct visual evidence are insufficient.  This includes both the state-of-the-art algorithm for answer grounding (i.e., MAC-Caps) and the mainstream VQA dataset~\cite{balanced_vqa_v2} which was designed to encourage models to look at the images.  

Our results reveal that the best indicator of better answer groundings is that models were pre-trained on the VizWiz-VQA dataset.  In particular, the top two approaches are LXMERT and MAC-Caps trained on VizWiz-VQA.  Interestingly, neither of these models were trained with answer groundings and so neither could benefit from direct supervision of what answer groundings look like.  Altogether, this finding highlights a considerable domain shift between the real-world use case for people with visual impairments and the contrived settings for generating the other datasets.

We also observe that, for the VQA task, the state-of-the-art VQA algorithms considerably outperform the state-of-the-art answer grounding models that are pretrained on the same datasets.  Specifically, compared to MAC-Caps, LXMERT predicts over 2.5 times as many correct answers while OSCAR predicts almost 2 times as many correct answers.  This finding suggests that a large part of the success for state-of-art VQA models still stems from learned biases that are unrelated to the relevant visual evidence.

In what follows, we conduct fine-grained analysis for the top-performing visual grounding model (MAC-Caps pre-trained on VizWiz-VQA) and the two state-of-the-art VQA models (LXMERT and OSCAR).

\vspace{-0.75em}\paragraph{Analysis With Respect to Image Coverage.}
\label{section:AlgorithmimgCoverage}
We next assess each model's ability to accurately locate the answer groundings based on the answer groundings' relative size. To do so, we divide the test VQA instances into three buckets of ``Small", ``Medium", and ``Large" based on their image coverage, specifically whether they occupy up to 1/3 of the image, between 1/3 and 2/3 of the image, and more than 2/3 of the image respectively.  In total, there are 1456 small examples, 458 medium examples, and 459 large examples in the test set.  Results are shown in Figure \ref{fig:AlgorithmImageCoverage}. 

Overall, we observe that all models struggle most to predict answer groundings for the small set. While this finding is not necessarily surprising, we believe it is still worthwhile to be shown experimentally.

Interestingly, the advantage the top-performing answer grounding model (MAC-Caps) has over the two state-of-the-art VQA models (LXMERT and OSCAR) in grounding answers stems from its ability to better ground small and medium sized regions.  This is evident when examining the median IoU scores, which are roughly double for MAC-Caps what is observed for the other two models.  A valuable area for future work will be to decipher what enables MAC-Caps to better attend to these smaller answer regions compared to the mainstream VQA models.

\vspace{-0.75em}\paragraph{Analysis With Respect to Image Quality.}
We next assess each model's ability to accurately locate the answer groundings based on the image quality issues defined in \cite{chiu2020assessing}.  First, we follow the process described in \cite{gurari2020captioning} to divide the VQAs into three buckets of ``None", ``Medium", and ``Severe" quality issues based on quality ratings from five crowdworkers, resulting in 1,930, 351, and 92 examples respectively.  Results are shown in Figure \ref{fig:AlgorithmImageQuality}.  We also assess performance with respect to specific quality issues (i.e., poor framing, blurry, too dark, too bright, obfuscations, and improper rotations) and, due to space constraints, provide results in the Supplementary Materials. 

Surprisingly, we observe that images without any quality issues (``None") are the most challenging for models, particularly the VQA models (LXMERT and OSCAR). Upon further analysis, we found that this is because these images have smaller answer regions, and hence show similar performance as observed for images with ``Small" attention areas for image coverage. Specifically, the average attention areas for ``None", ``Medium" and ``Severe" images are 0.28, 0.47 and 0.83 respectively. This also explains why models performed better on ``Severe" images as they tend to have larger answer regions. Further analysis also shows 82\% ``Severe" images contain questions asking about color, which we will see in the next section, are also examples where the models typically perform better.  


\vspace{-0.75em}\paragraph{Analysis With Respect to Vision Skills.}
Next, we assess each model's ability to accurately locate the answer groundings based on the vision skills needed to answer the questions, as introduced in Section~\ref{section:visionskill}. Specifically, these skills are object recognition, color recognition, text recognition, and counting.  Results are reported in Figure \ref{fig:AlgorithmSkill}. 

We consistently observe across all models that they perform worse for questions involving text recognition and counting while they perform better for questions involving object recognition and color recognition.  We suspect we observe improved performance for the latter two skills because color recognition has a relatively simple image analysis component and object recognition models have become quite advanced compared to many other vision tasks, given the large focus on the problem which was spurred by the ImageNet dataset challenge~\cite{deng2009imagenet}.  This finding also could be due in part to our prior observation that answer groundings for text recognition tend to have smaller answer groundings while color recognition tends to have larger visual grounding areas (Section~\ref{section:dataAnalysis}) and models perform worse for smaller answer groundings  (Section~\ref{section:AlgorithmimgCoverage}).

\begin{figure}[t!]
     \centering
     \includegraphics[width=0.6\textwidth]{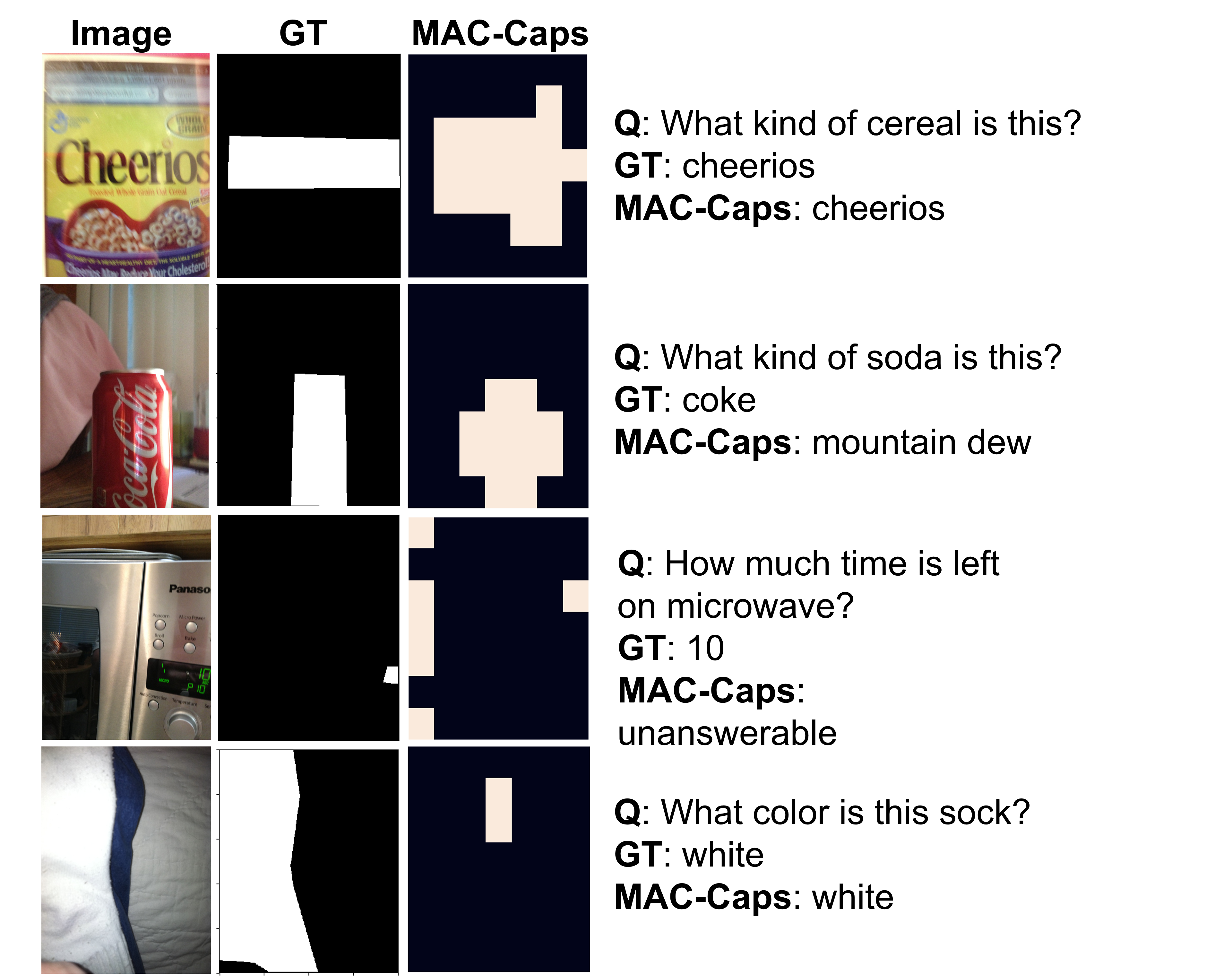}
     \vspace{-1.5em}
        \caption{Qualitative results exemplifying answer groundings from the best model, MAC-Caps pretrained on VizWiz-VQA. 
        }
    \label{fig:qualitative-results}
\end{figure}

\vspace{-0.75em}\paragraph{Analysis of the Presence of Text.}
We also analyze how often answer groundings contain text. We apply the Microsoft Read API to detect and recognize text in grounded areas.  Of the 9,998 answer groundings, about 52\% (i.e., 5,207 images) were detected as containing text.  For this subset, we compared the extracted text to the ground truth answers. We found that the ground truth answer was present in the detected text for only 7\% (i.e., 372 images) of visual questions.  This suggests that most language answers cannot be found directly from detected text in answer groundings.  

\vspace{-0.75em}\paragraph{Qualitative Results.}
We finally show qualitative results from the top-performing answer grounding model, MAC-Caps pretrained on VizWiz-VQA.  Results are shown in Figure~\ref{fig:qualitative-results}.  These examples illustrate answer groundings for a large range of sizes (e.g., row 3 vs row 4) as well as visual questions that require different vision skills, such as text recognition for rows 1 and 3, object recognition for row 2, and color recognition for row 4.  When observing the model's predictions, we observe that the results reinforce our quantitative findings that the model often fails.  These answer grounding failures include for both when the model predicts correct natural language answers (rows 1 and 4) and predicts incorrect answers (rows 2 and 3).

\section{Conclusions}
Our VizWiz-VQA-Grounding dataset offers a strong foundation for supporting the community to design less biased VQA models and more accurate answer grounding models which can serve as a valuable precursor for a range of practical applications.  We will publicly-release the dataset alongside a public evaluation server and leaderboard to spur community progress on this important answer grounding problem.  Our benchmarking of state-of-the-art models reveal current limitations for future models to overcome.  Future work will need to establish how to ensure such algorithms truly learn to understand the visual questions rather than learning superficial correlations between properties of visual questions and their answer groundings. 

\paragraph{Acknowledgments.}
We are grateful for funding support from Microsoft AI4A and Amazon Mechanical Turk.  %
\clearpage
{\small
\bibliographystyle{ieee_fullname}
\bibliography{main}
}
\clearpage

\section*{Supplementary Material}
This document supplements the main paper with more information about:
\begin{enumerate}
      \item Method for filtering visual questions (supplements Section 3.1)       
      \item Dataset collection (supplements Section 3.1) 
    
    \begin{itemize}
            \item Method for hiring expert crowdworkers (supplements Section 3.1)
            \item Screenshot of our annotation task interface (supplements Section 3.1)
            \item Method for reviewing work from crowdworkers (supplements Section 3.1)
    \end{itemize}
    
    \item Dataset analysis (supplements Section 3.2)
    \item Qualitative results for model benchmarking (supplements Section 4) 
\end{enumerate}

\renewcommand\thesection{\Roman{section}}
\setcounter{section}{0}
\section{Method for filtering visual questions}\label{section:methodOfFiltering}
Our aim was to ensure our dataset focused on visual questions for which answers could be unambiguously grounded to a single region. To do so, we performed five rounds of filtering visual questions from the initial VizWiz-VQA dataset, that we decribe below. 

First, we applied a filter to \textit{remove all questions which were unanswerable}.  This occurs regularly in the VizWiz-VQA dataset because the photographers could not verify the content in their images due to being blind.  We removed all visual questions which were labelled as ``unanswerable" in the provided ``answer\_type" metadata. 

We also removed visual \textit{questions for which at least half of the crowd did not agree on the same answer}.  We used a stringent string matching approach to detect if at least 5 out of the 10 answers per visual question are identical.  

Another filter we applied is to remove all so-called \textit{``questions" that actually embed multiple questions}.  An example is ``What is this and is it to be put in the microwave? Or does it even say?"  We did not simply filter visual questions with more than two question marks because we observed that users often ask refinements to their questions or ask the same question several times in different ways. Examples include ``What color are these jeans? Pink or gold?" and ``Can I wear these two pieces together? Do they match?".  From pilot testing, we observed an effective automated mechanism is to remove visual questions that have more than five words while containing the word ``and" (i.e., 901 visual questions).  We also process those visual questions that contained repetition of a single question, e.g., ``what is this? what is this?". We trim these down to a single question to remove redundancy.   We then had two expert crowdworkers review each candidate visual question to identify whether each question actually asked more than one question.  We filtered a visual question if at least one person flagged it as containing more than one question.  Of the 11,085 visual questions that were reviewed, 0.6\% (66) were tagged as containing more than one question. 

\begin{figure*}[t!]
     \centering
     \includegraphics[width=0.5\textwidth]{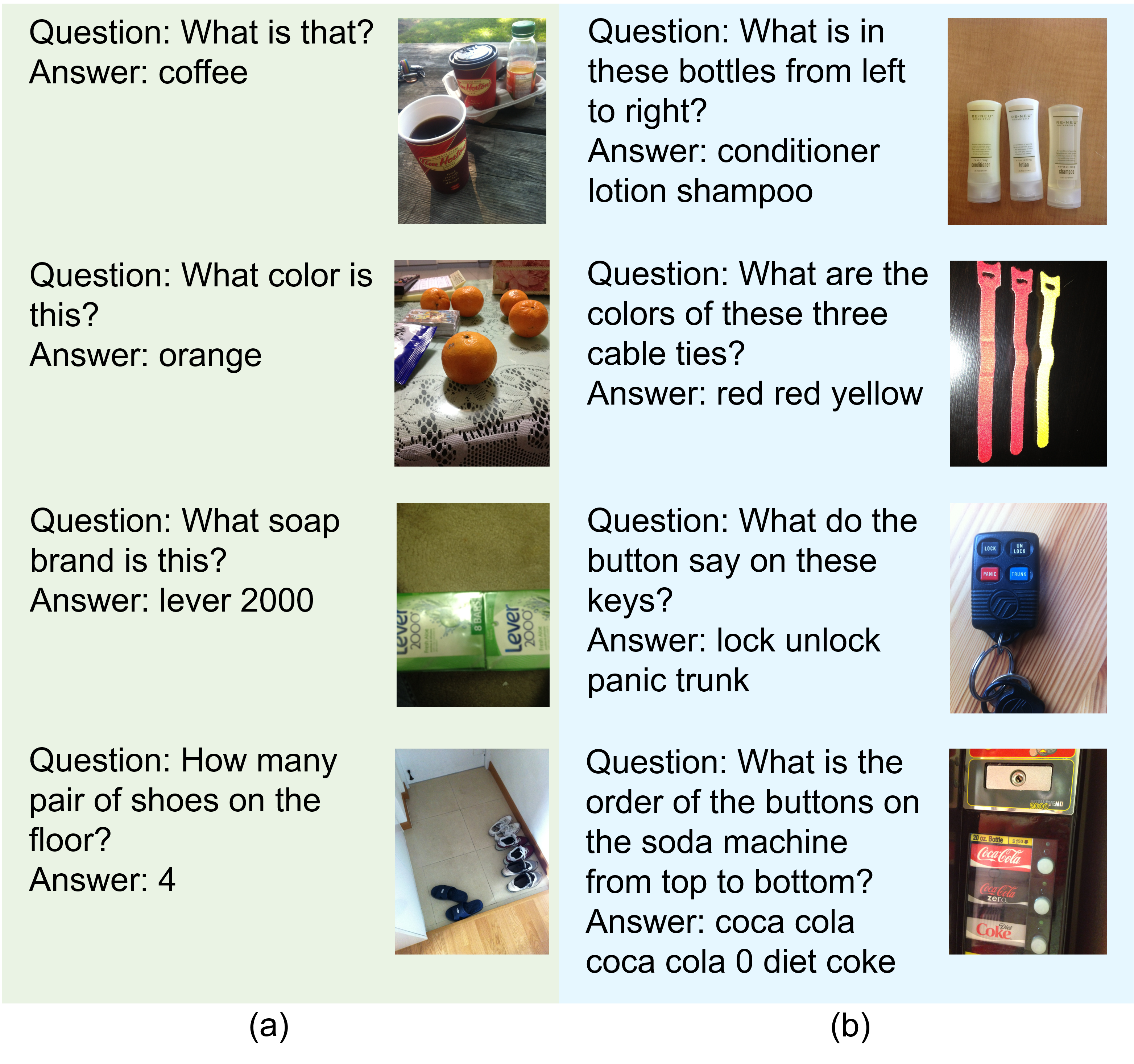}
        \caption{Answer groundings that refer to multiple regions. }
    \label{fig:instancesMultipleRegions}
\end{figure*}

We also filtered visual \textit{questions for which there is ambiguity where the answer is located in an image}, meaning answers referred to more than one image region.  We conducted preliminary analysis to understand the prevalence of this case. From 200 random visual questions from the VizWiz training dataset, we found only two contained this issue. From further analysis, we found that this issue often appears when the questions contains plural nouns\footnote{We used the NLTK package to detect parts of speech, including plural nouns.} or when they contain phrases such as ``how many". While the occurrences were rare, we still had two expert crowdworkers review each VQA to identify whether more than one polygon is needed to locate the region to which the answer is referring.  A VQA was removed if at least one person flagged it as needing more than one polygon.  Of 11,019 visual questions that were reviewed, 1.66\% (183) were marked as requiring more than one polygon to locate the answer grounding region.  From inspection of some of these visual questions, possible reasons for why an answer refers to more than one region are that multiple regions are suitable for arriving at the same answer (Figure~\ref{fig:instancesMultipleRegions}a) and that the answer actually embeds multiple answers that align with distinct visual content (Figure~\ref{fig:instancesMultipleRegions}b). 

Finally, we filtered visual \textit{questions that were answerable but could not be grounded}.  For example, often this occurs for questions that lead to the answer ``No", such as ``Is there a key?".  To do so, we had two expert crowdworkers review each VQA triplet to identify whether the answer is not shown in the image.  A VQA was removed if at least one person flagged it as not present.

\section{Dataset Collection}
\subsection*{Method for hiring expert crowdworkers}
We only accepted candidates who previously had completed at least 500 Human Intelligence Tasks (HITs) with over a 95\% acceptance rate and were from the United States.  The latter requirement gave us some confidence that the workers had English proficiency.  

Next, we required the crowdworkers to pass a qualification test which included challenging grounding tasks covered in our instructions.  In this test, each worker is asked to annotate 10 QA pairs for which the ground truth (GT) annotations were manually drawn by us. The grounding annotation by the worker is deemed correct if the region has more than a 70\% IoU score with the GT region. The workers had to annotate all of the 10 QA pairs correctly to pass the qualification test. The user interface blocked a worker from moving to the next of the 10 tasks until the generated annotation sufficiently matched our pre-annotated ground truth.   Completing this qualification task ensured a worker understood the task and how to handle challenging annotation scenarios.  

All workers who passed the qualification test were eligible to complete 20 grounding tasks.  In total, 27 workers completed these tasks.  We reviewed all annotations from them and hired nine workers who consistently generated high quality results. We limited our number of workers because we prioritized high quality annotations more than the efficiency from having more workers; i.e., it is easier to track the performance of fewer workers.

\subsection*{Screenshot of our annotation task interface}
\begin{figure*}[t!]
\begin{center}\includegraphics [scale=0.75] {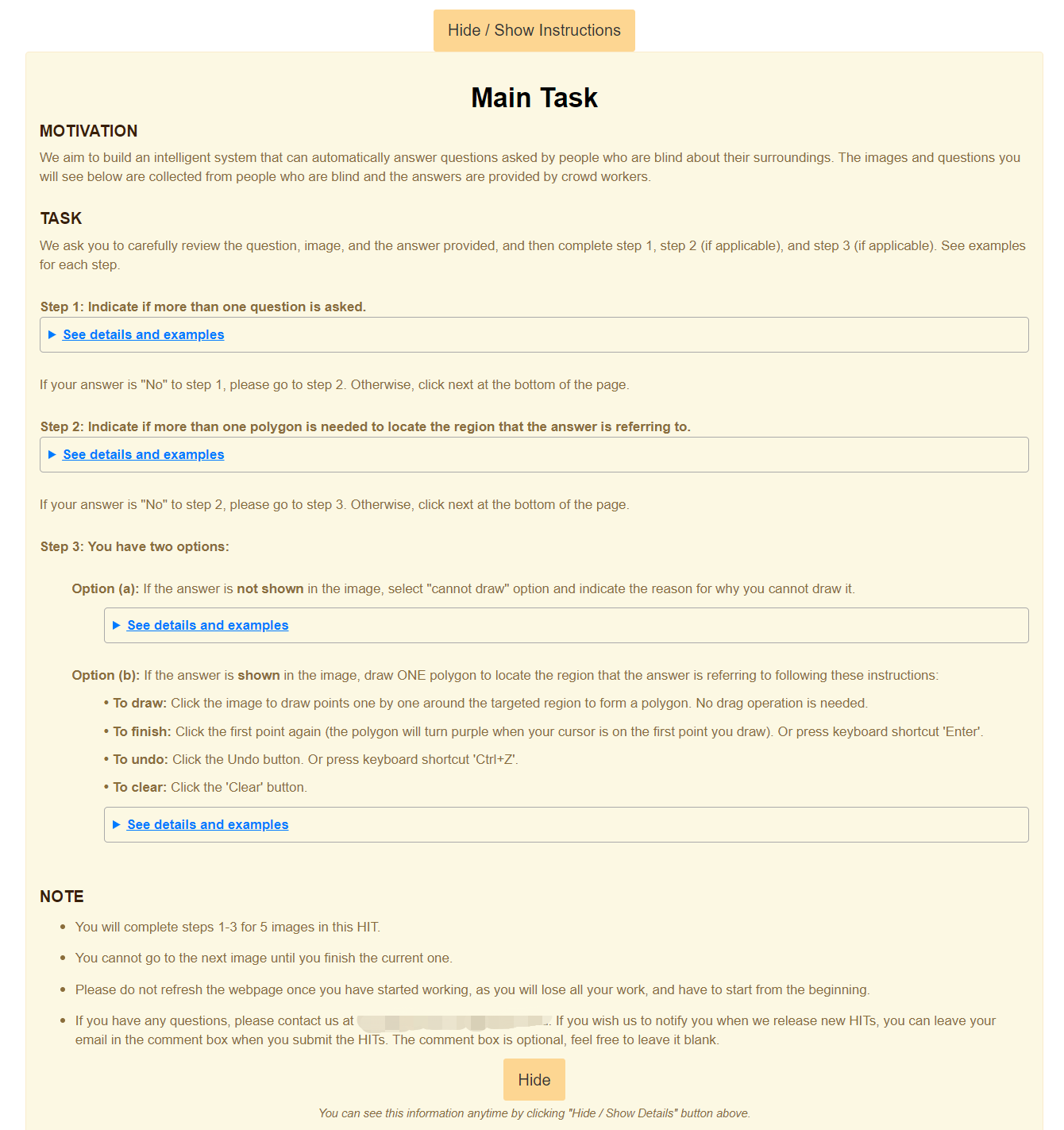}

\end{center}
  \vspace{-1.25em}
  \caption{Instructions for our annotation task interface.}
  \label{fig:instruction1}
\end{figure*}

\begin{figure*}[t!]
\begin{center}\includegraphics [scale=0.6] {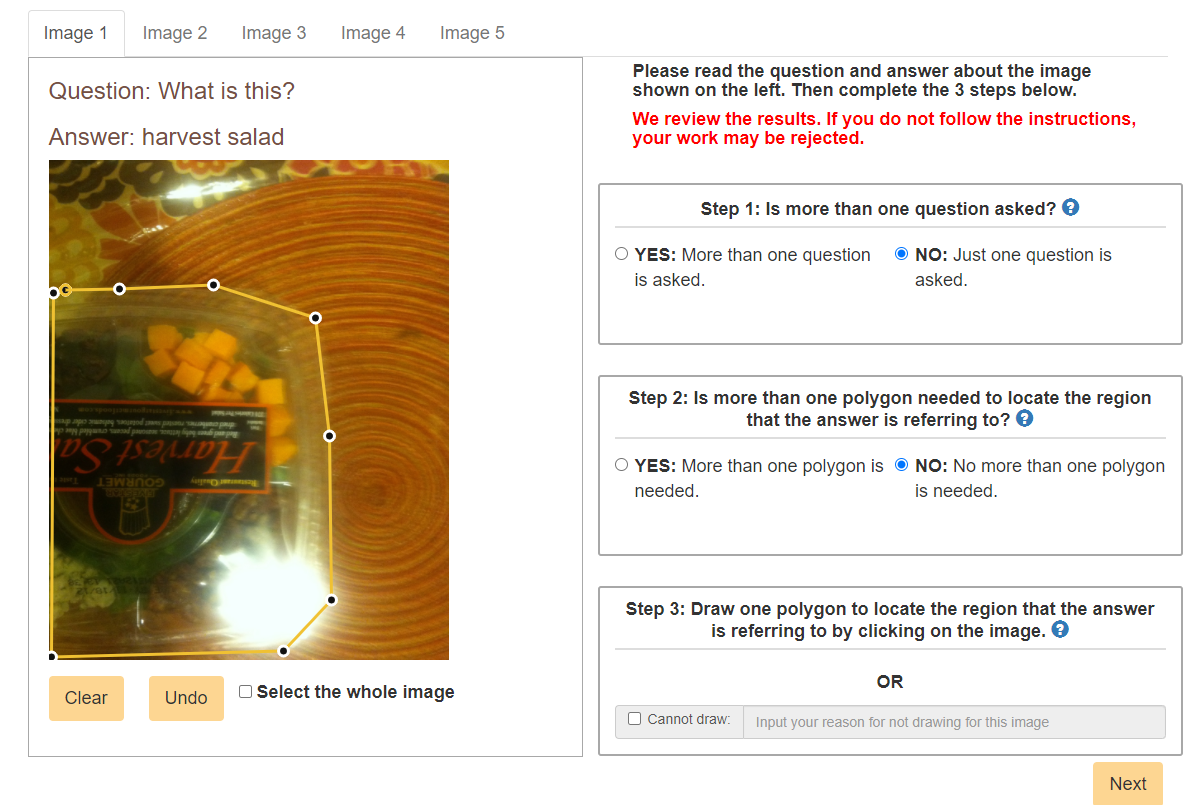}
\end{center}
    \vspace{-2.5em}
  \caption{A screenshot of our annotation task interface.}
  \label{fig:instruction2}
\end{figure*}

We show the crowdsourcing task that we created to collect grounding annotations, including a screen shot of the instructions in Figure \ref{fig:instruction1} and the annotation task in Figure \ref{fig:instruction2}. The link to this code is available at  \url{https://github.com/CCYChongyanChen/VizWizVQAGroundingCrowdSourcing}.

\subsection*{Method for reviewing work from crowdworkers}
\begin{figure*}[!t]
    \centering
    \includegraphics [scale=0.35] {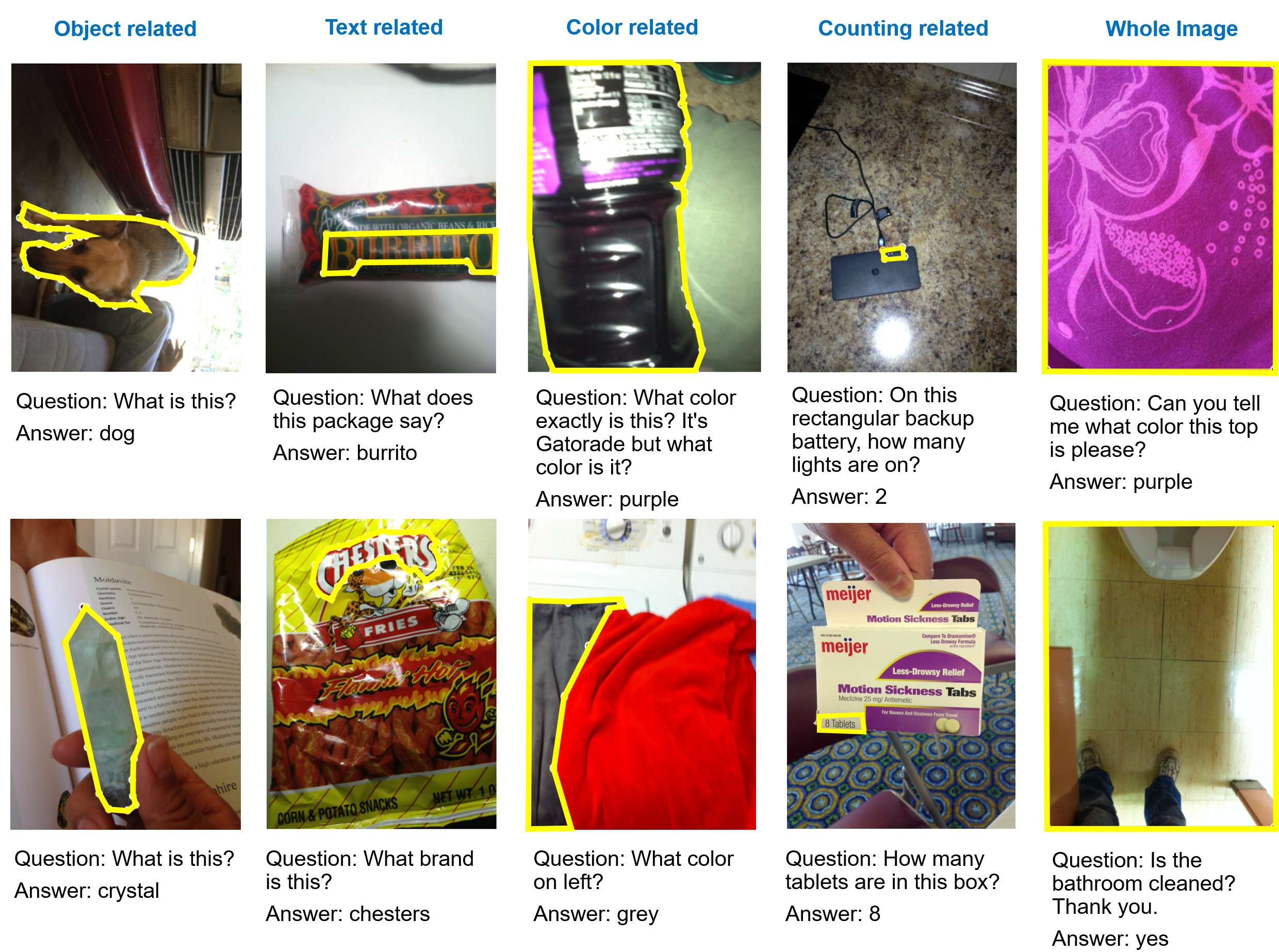}
 \vspace{-1em}
 \caption{Examples of answer groundings for a variety of question types.}
  \label{fig:examplesdifferentquestions}
\end{figure*}

The nine workers hired to create all answer groundings were given our contact information so they could contact us with any questions and we gave them the link to a live document where we frequently added our feedback to their questions about tricky examples.  As they submitted their work, we leveraged automated quality control checks and manually inspected random samples of their results to ensure the annotation quality remained high.  For the automated checks, we used the following rules to help us identify quality issues.  For each HIT, we recorded the total number of times a worker answered `Yes' in Step 1 (contains multiple questions), `Yes' in Step 2 (contains multiple regions), and `cannot draw' in Step 3 (rather than drawing a segmentation). If the total number for any of these steps was more than 3, we reviewed all results in the HIT. This is because, our pre-processing steps (described in Section \ref{section:methodOfFiltering}) meant that most of the QA pairs should contain one question and our preliminary analysis indicated that we seldom observed either that there are multiple regions or that the answer is not present in the image.  We also monitored the time the worker spent on each HIT. If it was less than 30 seconds, we inspected the results.\footnote{In pilot studies, we found that initially crowdworkers took an average of 4.38 mins to finish a HIT but this time dropped to 2.8 mins as workers became familiar with the task.}  Finally, we also calculated the number of points used to draw the polygon. If a worker drew less than 5 points for an answer grounding for more than two visual questions in a single HIT, we manually inspected all results for the HIT.  Examples of high quality answer grounding results are shown in Figure \ref{fig:examplesdifferentquestions}. 


\clearpage
\clearpage
\section{Dataset Analysis}

\begin{figure}[h!]
    \centering
    \includegraphics [width=0.38\textwidth]{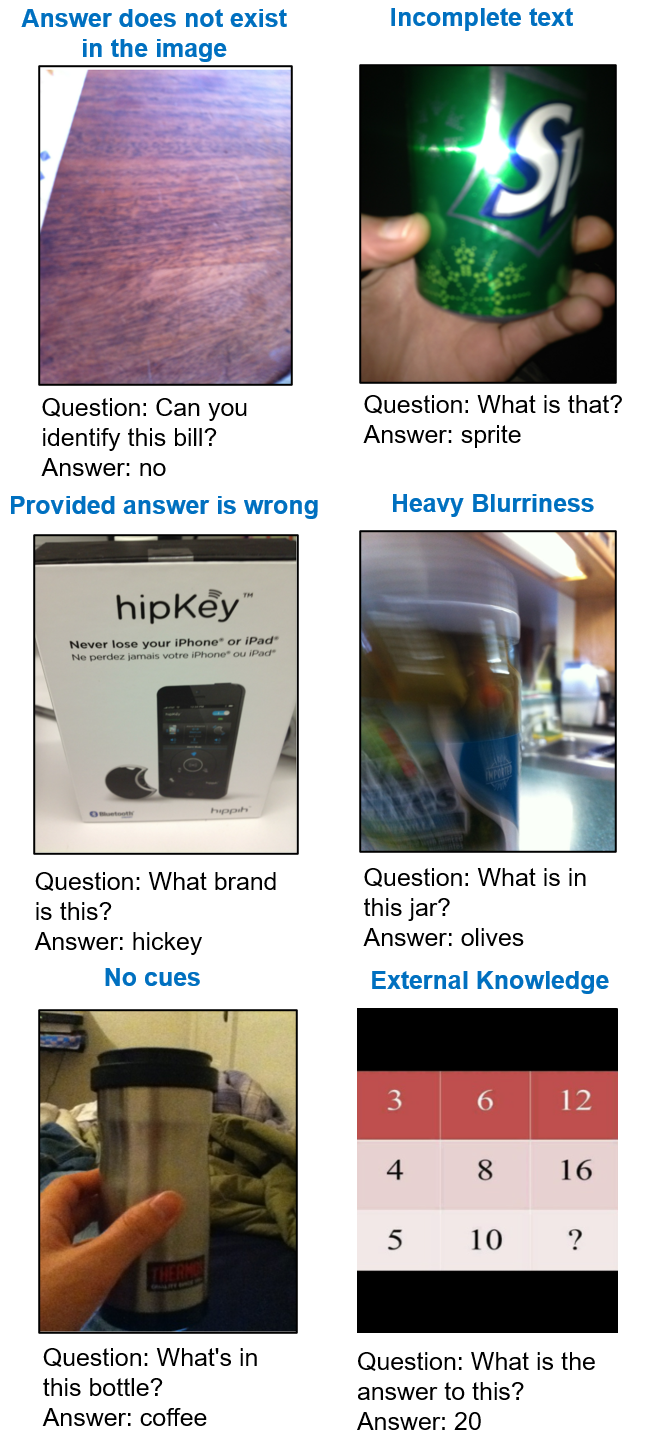}
    \caption{Examples of visual questions for which workers indicated the answer cannot be grounded in the image paired with a more general reason why (shown in blue text). }
    \label{fig:nodrawexample}
\end{figure}

\paragraph{Reasons why crowdworkers selected ``cannot draw".}
The crowdworkers explained why they indicated the answer cannot be localized in the image (i.e., by selecting ``cannot draw" in the task user interface) in a free-text box.  The top 10 reasons are `nothing to draw a polygon' (106 times), `incomplete text' (28 times), `wrong answer' (25 times), `the image is too blurry/blurry' (26 times), `no clue(s)' (20 times), `subjective' (9 times), `the answer is no so therefore can't draw something not there' (7 times), `nothing on screen' (4 times), `answer is not shown in the image and the answer can be answered without the image' (4 times), and `nothing found' (4 times). Examples of these flagged visual questions are shown in Figure \ref{fig:nodrawexample}.   

\begin{figure}[h!]
    \centering
    \includegraphics [width=0.4\textwidth]{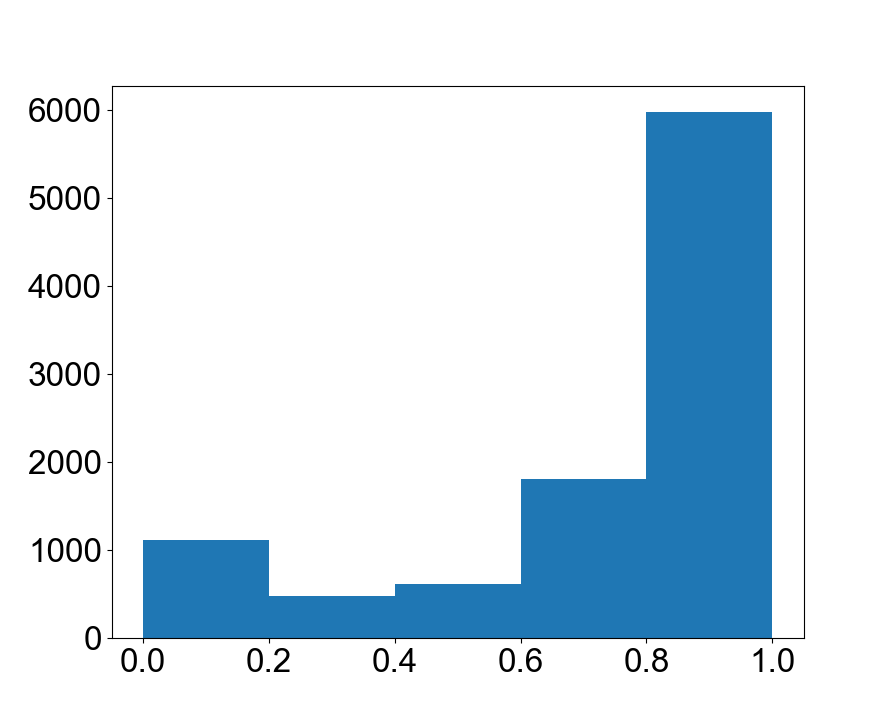}
    \caption{Histogram of IoU scores indicating similarity between each pair of answer groundings per visual question. The majority have high agreement, in the range between 0.8 and 1.0. }
    \label{fig:histogramIOU}
\end{figure}

\begin{figure}[h!]
    \includegraphics [scale=0.4]{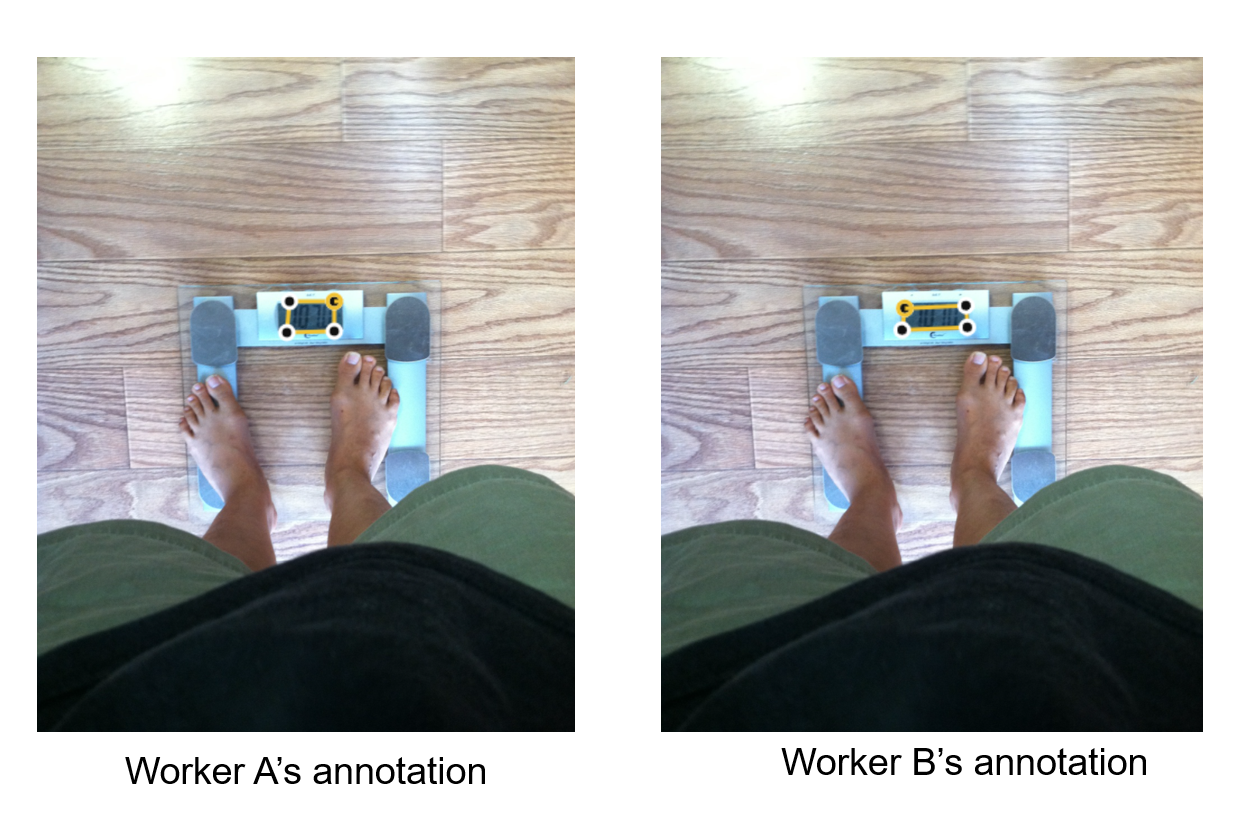}
    \caption{Example of two answer groundings for a small area. The IoU score between these two annotations is 58\%. This exemplifies the tendency for IoU scores to be low for small regions when groundings appear similar.}
    \label{fig:textIOU}
\end{figure}

\vspace{-0.75em}\paragraph{Annotation agreement. } 
Recall that two answer grounding annotations were collected per visual question from two crowdworkers.  A histogram showing the IoU scores between each pair of annotations per visual question across the 9,998 visual questions is shown in Figure \ref{fig:histogramIOU}.  The majority (around 60\%) of the IoU scores are between 0.8 and 1.0, around 20\% lie between 0.6 and 0.8, and slightly more than 10\% lie between 0 and 0.2. This shows that, typically, there is high annotation agreement. We attribute low scores in part to the IoU being a poor metric when accounting for smaller regions. An example is shown in Figure \ref{fig:textIOU}, where the IoU score is $\sim$58\% despite that both annotations are visually similar and correct.  We selected as ground truth the larger region from the two groundings since often the smaller one is contained in the larger one. 

\vspace{-0.75em}\paragraph{Whole Image}
The visual answer is labelled as referring the whole image for 0.9\% (903) of visual questions.  Often, workers selected ``Whole Image" when the question related to color, the camera is set too close to the object, or the questioner asks about the general description of the scene. Examples are shown in the last column of Figure \ref{fig:examplesdifferentquestions}.

\begin{figure}[h!]
\centering
    \includegraphics [scale=0.45]{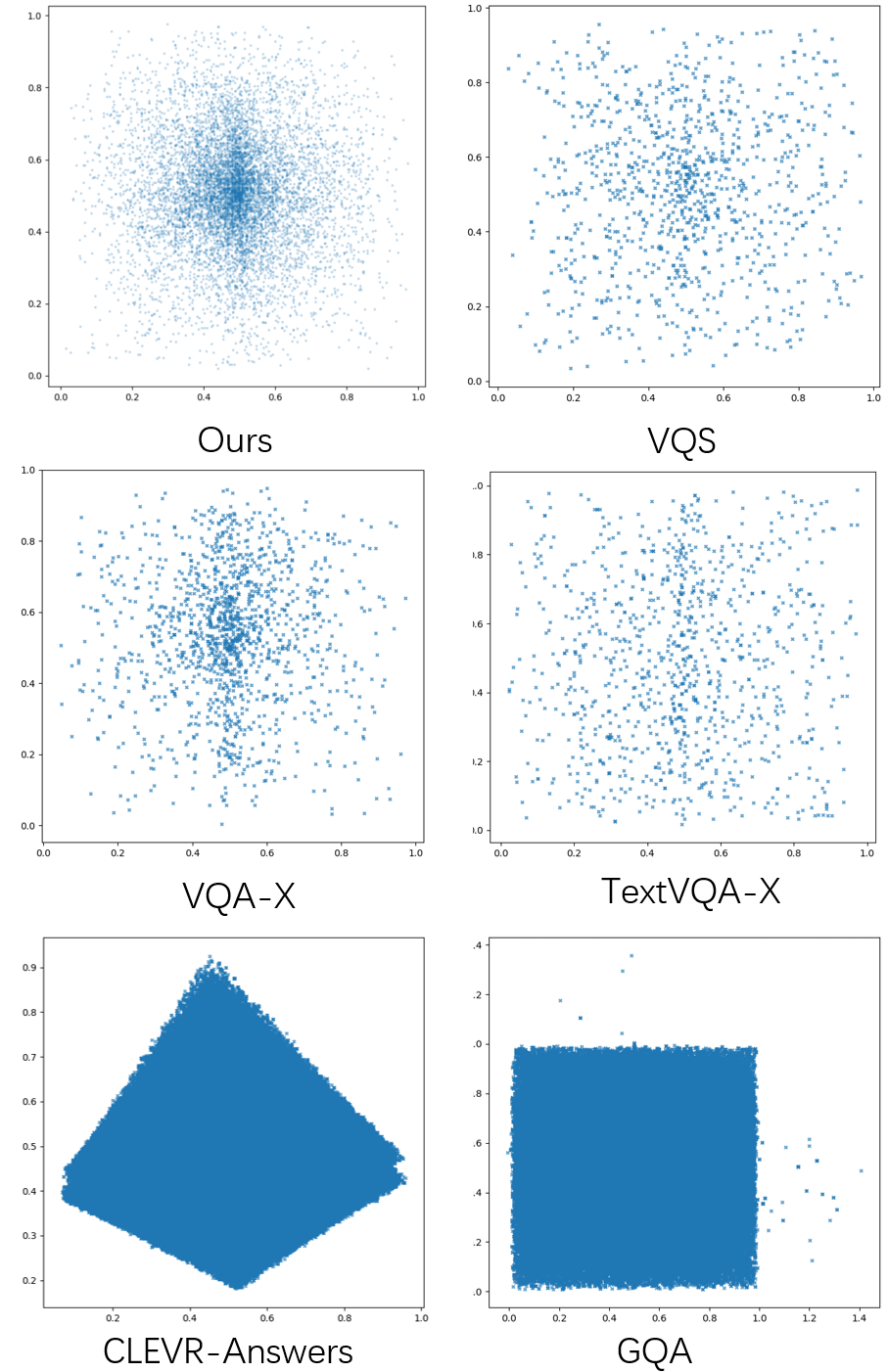}
    \caption{Center of the mass of answer groundings in the VizWiz-VQA-Grounding, VQS, VQA-X and the TextVQA-X datasets.}
    \label{fig:centerofmass}
\end{figure}

\vspace{-0.75em}\paragraph{Location of answer grounding. }
Expanding on Table 1 in the main paper, we visualize the center of mass for all answer groundings in the different datasets.  Results are shown in Figure \ref{fig:centerofmass}.  As shown, our new dataset clusters as a circle and has a smaller range of values than the other datasets.

\begin{figure}[h!]
\begin{center}\includegraphics [scale=0.4] {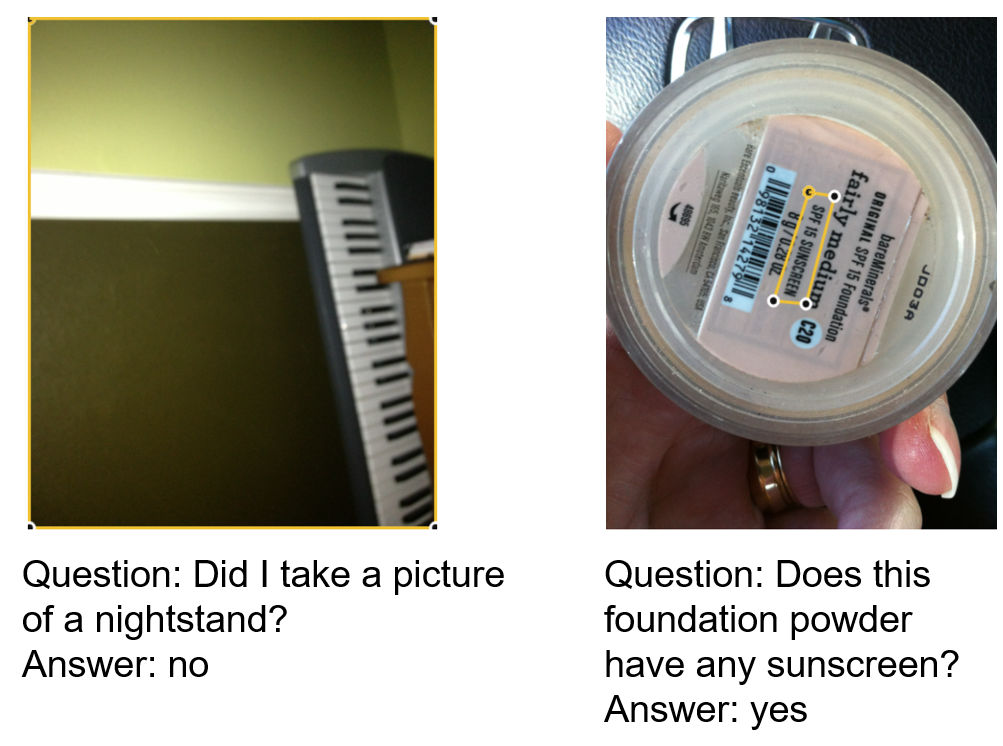}
\end{center}
\vspace{-1em}\caption{Example illustrating the trend that the `yes' answer groundings are typically smaller than the `no' answer groundings.}
\label{fig:yesno}
\end{figure}

\vspace{-0.75em}\paragraph{Grounding of most common answers.}
Expanding on the analysis in the main paper, we report statistics about the most common answers in Table \ref{tb:statCommonAnswer}. Intuitively, we observe that objects with rectangular shapes need fewer points, e.g., keyboard, laptop, and shampoo. In contrast, objects with complex boundary require more points, e.g., dog, hand, chair, and cat.  Less intuitively, the mean of the grounding size for the answer `yes' is smaller than those for which the answer is `no'. We suspect groundings for `no' answers more often refer to the whole image while `yes' primarily focus on specific regions, as exemplified in Figure \ref{fig:yesno}.

\begin{table}[]\centering
\begin{tabular}{l|l|l|l}
\hline
\textbf{Answer}                  & \textbf{Size (\%)}   & \textbf{Points} & \textbf{Images}  \\ \hline
\textbf{yes}  
 & 89,264 (45\%)&13 &525   \\ \hline
\textbf{no}            
 & 112,648 (56.3\%)&10 &240   \\ \hline
\textbf{keyboard} 
 &125,786 (62.9\%)& 7 &118  \\ \hline
\textbf{dog} 
 & 63,598 (31.8\%)& 40&85  \\ \hline
\textbf{laptop} 
 & 122,471 (61.2\%)& 9&66  \\ \hline
\textbf{pepsi} 
 & 55,055 (27.5\%)& 12&48  \\ \hline
\textbf{coca cola} 
 & 47,644 (23.8\%)& 11&46  \\ \hline
\textbf{orange} 
 & 82,201 (41.4\%)& 14 &42  \\ \hline
\textbf{corn} 
 & 57,947 (29.0\%)& 11&37  \\ \hline
\textbf{green beans} 
 & 56,246 (28.1\%)& 10&34  \\ \hline
\textbf{pen} 
 &16,731 (8.4\%)&17 &33  \\ \hline
\textbf{lotion} 
 &42,835 (21.4\%)& 16&32  \\ \hline
\textbf{cat} 
 & 54,714 (27.4\%) &34&30  \\ \hline
\textbf{water bottle} 
 & 61,062 (30.5\%)& 24 &28  \\ \hline
\textbf{phone} 
 & 98,399 (49.2\%) &15 &28  \\ \hline
\textbf{soup} 
 &44,875 (22.4\%)& 11&28  \\ \hline
\textbf{Shampoo} 
 & 35,190 (17.6\%)& 8&26  \\ \hline
\textbf{hand sanitizer} 
 & 68,540 (34.3\%)&23 &26  \\ \hline
\textbf{hand} 
 & 86,867 (43.4\%)& 39&26  \\ \hline
\textbf{chair} 
 &81,435 (40.7\%)&37 &26  \\ \hline
\textbf{remote} 
 & 63,090 (31.5\%)&17 &26  \\ \hline
\end{tabular}
\caption {Mean value of properties describing the regions for most common answers (excluded color-related answer, as it is reported in Table 2.}
\label{tb:statCommonAnswer}
\end{table}

\begin{figure*}[h!]
    \includegraphics [scale=0.95]{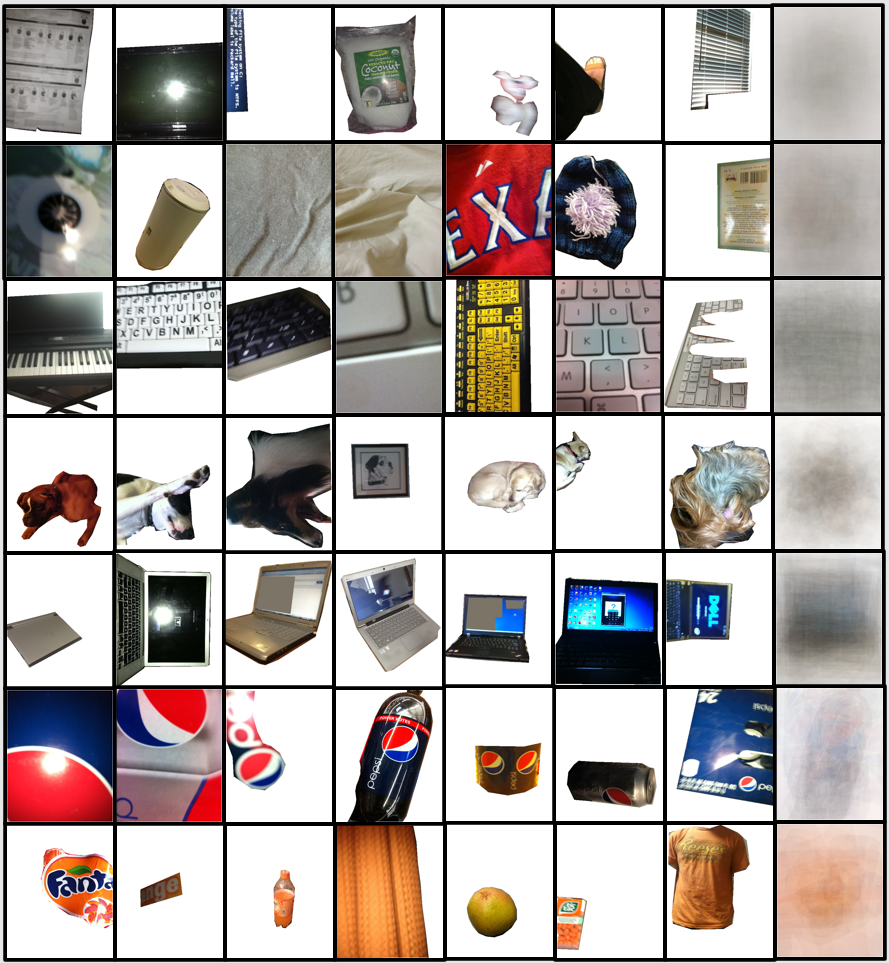}
    \caption{Examples of answer groundings in our VizWiz-VQA-Grounding dataset.  Shown are answering groundings for common answers (i.e., yes, no, keyboard, dog, laptop, pepsi, orange) as well as the average image across all groundings for each answer.}
    \label{fig:Avg_image}
\end{figure*}

\vspace{-1.5em}
We show for the most common answers examples of the grounded area as well as the average images in Figure \ref{fig:Avg_image}. For the same language answer, visual groundings can be diverse. For example, the dog has different breeds, colors, postures, and locations. The dog can be partially visible and under different illumination. Also, the answer ``dog" can refer to an animal or a picture of a dog. A more blurry/grey average image is indicative of a greater diversity of images for the answer.

\vspace{-0.75em}\paragraph{Dataset comparison.}
We summarize how the characteristics of the eight related answer grounding datasets, discussed in the related work section of the main paper, relate and differ to our dataset in Table~\ref{datasets}.  Our summary indicates the visual annotation type, number of images with visual annotation, and source VQA dataset.  

\begin{table*}[h!]
\begin{tabular}{l|l|l|l}
\hline
\textbf{Dataset}  & \textbf{Visual Annotation Type} & \textbf{\begin{tabular}[c]{@{}l@{}}\# Images \\ ($\times$ Annotations per Image)\end{tabular}}& \textbf{VQA dataset}\\ \hline

Visual7W (2016)\cite{Zhu_2016_CVPR} & Bounding box                     &        47,300    & COCO \cite{lin2014microsoft} \\ \hline
VQA-HAT (2017) \cite{das2017human}& Human Att. (deblur image)&   59,849          &    VQA v1     \\ \hline
VQS (2017) \cite{gan2017vqs}    & Segmentation+bounding box    & 37,868        &  COCO    \\ \hline
VQA-X (2018) \cite{huk2018multimodal}  & Segmentation                     &     6,000 ($\times$ 1)     & COCO, VQAv2    \\ \hline

GQA  (2019)   &   Bounding box           & 355,530 ($\times$1)      &  GQA (2019) \cite{hudson2019gqa}  \\ \hline
AiR (2020)  \cite{chen2020air}    & Human Att. (eye-tracking) &    987 ($\times$ 20)      &      GQA \cite{hudson2019gqa} \\ \hline
Text VQA-X (2021) \cite{nagaraj-rao-etal-2021-first}    &   Segmentation (brush)  &            11,681($\times$ 1) &     TextVQA\cite{singh2019towards}   \\ \hline
CLEVR-Ans (2021) \cite{urooj2021found}  &  Bounding box       &          445,268 ($\times$1)       &  CLEVR  \\ \hline
Ours    & Segmentation                     & 9,998 ($\times$ 2)         &     VizWiz-VQA \cite{gurari2018vizwiz}  \\ \hline
\end{tabular}
\vspace{-0.5em}
\caption{Comparison between existing VQA answer grounding datasets and our dataset. 
}
\label{datasets}
\end{table*}

\section{Algorithm}

\paragraph{mAP@IoU results.}
The performance for each model on the VizWiz-VQA-Grounding test split with respect to mAP@IoU score is shown in Table \ref{table:MAPIOUresults}. Overall, the low mAP scores reinforce our findings in the main paper that the models perform poorly on our new dataset and that the best indicator of better answer groundings is that models were pre-trained on the VizWiz-VQA dataset (Sec 4). 
\vspace{-0.5em}
\begin{table}[h!]
\small
\hspace{-1em}
\centering
\begin{tabular}{ccccc}
\hline
 \textbf{Model (Pretrained)} & \textbf{mAP25} & \textbf{mAP50} & \textbf{mAP75} & \textbf{mAP} \\
\hline
LXMERT  (VizWiz)   &  14.21\%           & 1.99\%          & 0.02\%           & 0.49\%                          \\
OSCAR   (VQA-v2)           &  7.10\%           & 0.15\%          & 0.00\%           & 0.03\%                          \\
 MAC-Caps (GQA)                 &  3.94\%           & 0.09\%          & 0.00\%           & 0.02\%                          \\
MAC-Caps (CLEVR)               &  6.38\%           & 0.12\%          & 0.00\%           & 0.01\%                          \\
MAC-Caps (VQA-v2)              & 9.32\%           & 0.45\%          & 0.01\%           & 0.08\%                          \\
MAC-Caps (VizWiz)          & \textbf{22.32\%} & \textbf{4.09\%} & \textbf{0.17\%}  & \textbf{0.96\%} \\
\hline
\end{tabular}
\caption{Performance of six models when evaluated on the VizWiz-VQA-Grounding test set: two state-of-art VQA models (LXMERT~\cite{tan2019lxmert} and OSCAR~\cite{li2020oscar}) and four variants of the state-of-art VQA model for answer grounding (MAC-Caps~\cite{urooj2021found}) with respect to mAP@IoU.  Results are provided based on the COCO evaluation protocol of using different IoU thresholds, from 0.25 to 0.75, and averaging AP values with IoU threshold ranges from 0.5 to 0.95 with a step size of 0.05} 

\label{table:MAPIOUresults}
\end{table}
\paragraph{Extraction of attention maps: baseline models.}
As discussed in the main paper, we extract attention maps for the two VQA models: LXMERT and OSCAR.  For LXMERT, the output consists of four different attentions: self-vision attention, self-language attention, image-guided question attention, and question-guided image attention. Following the authors' recommendation, we picked the  question-based image attention. For each visual question, the model predicts 12 attention heads, where each head was of size 20x36. We first average the attention maps for each head by length, and then over all the heads to obtain the final attention map.  For OSCAR, we extract attention weights from the last layer. It has 114x114 for 16 heads; For the 114 sequence length, the first 64 dimensions are the language attention, while the later 50 dimensions are image attentions. We uses self-vision attention.

\paragraph{Naive baseline.}
As a naive baseline, we treated predicting the whole image as the grounded area. This baseline receives an average IoU score of 33\%.

\begin{figure}[h!]
\hspace{-15em}
\begin{center}\includegraphics [scale=0.5] {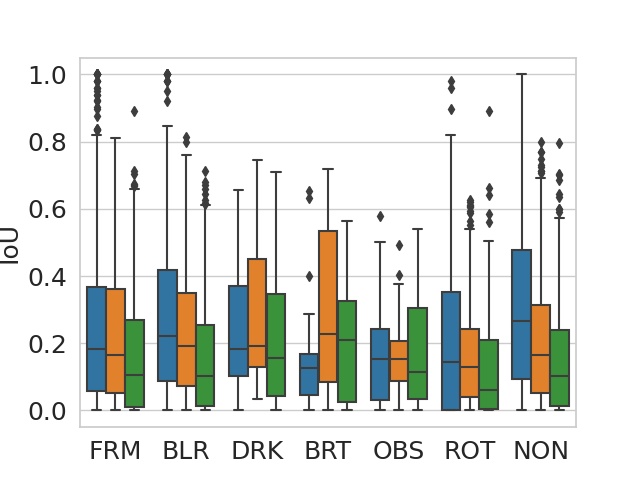}
\end{center}
\vspace{-0.25em}\caption{Comparison of MAC-Caps (pretrained on VizWiz), LXMERT, and OSCAR's performance on visual questions for images with different quality issues.}
\label{fig:AlgorithmImageQuality2}
\end{figure}

\vspace{-0.75em}\paragraph{Analysis With Respect to Image Quality.}
As mentioned in the main paper, we report here our fine-grained analysis to assess each model's ability to accurately locate the answer groundings based on the image quality issues defined in \cite{chiu2020assessing}: poor framing, blurry, too dark, too bright, obfuscations, and improper rotations. Results are shown in Figure \ref{fig:AlgorithmImageQuality2}.  We observe that images with obscured image quality issue are the most challenging to ground and images with the rotation issue are the second most challenging to ground. 

\vspace{-0.75em}\paragraph{Qualitative results.}
\begin{figure*}
     \centering
     \includegraphics[width=0.94\textwidth]{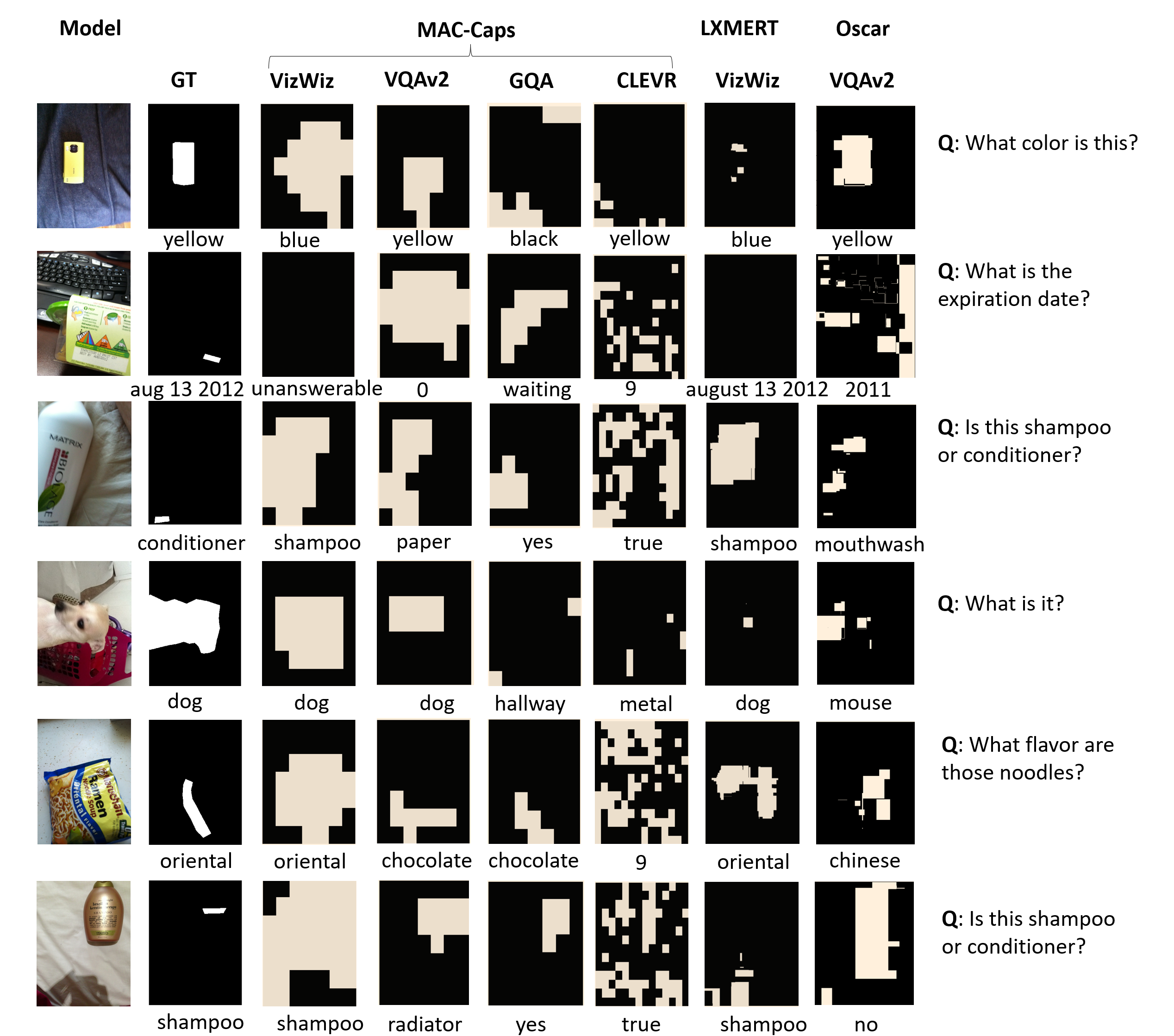}
     \hspace{-5em}
     \vspace{-0.5em}
        \caption{Qualitative results exemplifying answer groundings from 6 models. }
    \label{fig:qualitative-results-supp}
\end{figure*}
We show examples for the answer groundings predicted by the six benchmarked models.  Results are shown in Figure \ref{fig:qualitative-results-supp}. The third column exemplifies the top-performing MAC-Caps model pre-trained on VizWiz.  We observe examples for which the MAC-Caps pretrained on VizWiz predicts incorrectly (rows 1-3) and correctly (rows 4-6). Examples highlight that the MAC-Caps model struggles with images containing text (row 2, 3), can correctly predict the answer without grounding the correct region (row 6), and can fail to predict both the correct answer and answer grounding (row 2, 3).  We observe for LXMERT that it can similarly predict the answer and answering grounding incorrectly (column 6; row 1, 3) as well as predict the answer correctly while failing to accurately locate the answer grounding (column 7; row 2, 4, 5, 6).  We observe for the Oscar model pretrained on VQAv2 that it can detect the foreground object better compared to other models, but fails when the visual evidence is a small region (column 8; row 2, 3). 

We found that models (MAC-CAPs and LXMERT) pretrained on VizWiz-VQA failed to answer the question and locate the visual evidence for visual questions requiring color recognition (row 2, column 3,7), while models (MAC-CAPs and OSCAR) pretrained on VQAv2 can answer questions related to color and ground the correct region (row 1, column 4, 8).  We found this surprising since the VQAv2 has less visual questions related to color than the VizWiz dataset and also represents a distinct domain \cite{zengvision}.

\end{document}